\documentclass[article]{elsarticle}

\usepackage{lineno,hyperref}

\usepackage{amssymb}
\usepackage{graphicx}
\usepackage[margin=1.1in]{geometry}
\usepackage{float}
\usepackage{multicol}
\usepackage{color}
\usepackage{tabularx}
\usepackage{multirow}
\usepackage{caption}
\usepackage{algorithmicx}
\usepackage{algorithm}
\usepackage{algpseudocode}
\usepackage{multirow}
\usepackage{subcaption}
\usepackage{lineno}
\usepackage[T1]{fontenc}
\usepackage[latin1]{inputenc}
\usepackage{amsmath}
\usepackage{amsfonts}
\usepackage{booktabs}
\usepackage{siunitx}
\usepackage{placeins}

\begin{document}

\begin{frontmatter}

\title{A Nonlinear Orthogonal Non-Negative Matrix Factorization Approach to Subspace Clustering}

\author{Dijana Toli\'c \fnref{myfootnote}}
\address{Laboratory for Machine Learning and Knowledge Representations, Ruder Bo{\v s}kovi\'c Institute, Zagreb, Croatia, Bijenicka cesta 54, 10000 Zagreb, Croatia}
\fntext[myfootnote]{Corresponding author - email: dijana.tolic@irb.hr, tel: +385 1 457 1352}

\author{Nino Antulov-Fantulin}
\address{ETH Z\"urich, Swiss Federal Institute of Technology, Clausiusstrasse 50, COSS, CLU, 8092 Z\"urich, Switzerland}

\author{Ivica Kopriva}
\address{Laboratory for Machine Learning and Knowledge Representations, Ruder Bo{\v s}kovi\'c Institute, Zagreb, Croatia, Bijenicka cesta 54, 10000 Zagreb, Croatia}

\begin{abstract}
A recent theoretical analysis shows the equivalence between non-negative matrix factorization (NMF) and spectral clustering based approach to subspace clustering. As NMF and many of its variants are essentially linear, we introduce a nonlinear NMF with explicit orthogonality and derive general kernel-based orthogonal multiplicative update rules to solve the subspace clustering problem. In nonlinear orthogonal NMF framework, we propose two subspace clustering algorithms, named kernel-based non-negative subspace clustering KNSC-Ncut and KNSC-Rcut and establish their connection with spectral normalized cut and ratio cut clustering. We further extend the nonlinear orthogonal NMF framework and introduce a graph regularization to obtain a factorization that respects a local geometric structure of the data after the nonlinear mapping. The proposed NMF-based approach to subspace clustering takes into account the nonlinear nature of the manifold, as well as its intrinsic local geometry, which considerably improves the clustering performance when compared to the several recently proposed state-of-the-art methods.

\end{abstract}

\begin{keyword}
subspace clustering, non-negative matrix factorization, orthogonality, kernels, graph regularization

\end{keyword}

\end{frontmatter}


Introduced in \cite{20} as a parts-based low-rank representation of the original data matrix, non-negative matrix factorization (NMF) has shown to be a useful decomposition of multivariate data \cite{9,24,25}. 
The most important feature of NMF is the non-negativity of all elements of the matrices involved, which allows an additive parts-based decomposition of the data. This non-negativity is often encountered in real world data, providing a natural interpretation in contrast to other decomposition techniques that allow negative combinations (such as SVD). Related NMF factorizations include convex NMF, orthogonal NMF and kernel NMF \cite{31,30,50,32, 36, 101}.


The key idea in subspace clustering is to construct a weighted affinity graph from the initial data set, such that each node represents a data point and each weighted edge represents the similarity based on distance between each pair of points (e.g. the Euclidean distance). Spectral clustering then finds the cluster membership of the data points by using the spectrum of an
affinity graph. Spectral clustering can be seen as a graph partition problem and solved by the eigenvalue decomposition of the graph Laplacian matrix \cite{3,4, 15, 49, 48}.
In particular, there is a close relationship between the eigenvector corresponding to the second eigenvalue of the Laplacian and the graph cut problem \cite{7,8}. However, the complexity of optimizing graph cut objective function is high, e.g. the optimization of the normalized cut (Ncut) is known to be an NP-hard problem \cite{31,51,2, KoprivaCut}. Spectral clustering seeks to get the relaxed solution, which is an approximate solution for the graph partition. Compared with conventional clustering algorithms, spectral clustering has advantages to converge to global optimum and performs well for the sample space of arbitrary shape \cite{2,3,4,5}.

Despite empirical success of spectral clustering, one drawback is that a mixed-signed result given by the eigenvalue decomposition of the Laplacian may lack clustering interpretability or degrade the clustering performance \cite{9}. The computational complexity of the eigenvalue decomposition is $\mathcal{O}$(\emph{n}\textsuperscript{3}), where \emph{n} denotes the number of points. To avoid the computation of eigenvalues and eigenvectors, a recently established connection of the spectral clustering and non-negative matrix factorization (NMF) was utilized in \cite{16,17} and \cite {18}. As pointed out in \cite{17}, the formulation of non-negative spectral clustering is motivated by practical reasons: (i) one can use the update algorithms of NMF to solve spectral clustering, and (ii) NMF framework can easily incorporate additional constraints to spectral clustering algorithms.

It was shown in \cite{17} that spectral clustering Ncut can be treated as a symmetric NMF problem of the graph affinity matrix constructed from the data matrix. Similary, it was also proven that the Rcut spectral clustering is equivalent to the symmetric NMF of the graph affinity matrix, introducing the non-negative Laplacian embedding (NLE) \cite{18}. Both results \cite{17, 18} only factorize the graph affinity matrix, imposing the assumption that the input data comes in as a matrix of pairwise similarities. The factorization of the graph affinity matrix was replaced with the factorization of the data matrix itself in \cite{16}, and including an additional global discriminative regularization term in \cite{19}.
However, both NMF-based NSC methods \cite{16, 19}, minimize data fidelity term in the linear input space.

In this paper we propose a nonlinear orthogonal NMF approach to subspace clustering. We establish an equivalence with spectral clustering and propose two non-negative spectral clustering algorithms, named \textit{kernel-based non-negative spectral clustering} KNSC-Ncut and KNSC-Rcut. 
To further explore the nonlinear orthogonal NMF framework, we also introduce a graph regularization term \cite{25} which captures the intrinsic local geometric structure in the nonlinear feature space. 
By preserving the geometric structure, the graph regularization term allows the factorization method to have more discriminating power for clustering data points sampled from a submanifold which lies in a higher dimensional ambient space \cite{25}. 

Recently, a similar connection between kernel PCA and spectral methods  has  been  shown in \cite{52, 3, 5, Alzate2006}. Our method gives an insight into the connection between kernel NMF and spectral methods, where the kernel matrix from multiplicative updates corresponds to the nonlinear graph affinity matrix in spectral clustering. Different from \cite{16,19,17,18}, our equivalence is established by directly factorizing the nonlineary mapped input data matrix. To the best of our knowledge, this is the first approach to non-negative spectral clustering that uses kernel orthogonal NMF.

By constraining the orthogonality of the clustering matrix during the nonlinear NMF updates, the cluster membership can be obtained directly from the orthogonal clustering matrix, avoding the need of usual $k$-means clustering \cite{16,17,18,19}. The proposed approach has a total run-time complexity of $\mathcal{O}(kn^2)$ for clustering $n$ data points to $k$ clusters, which is less than standard spectral clustering methods $\mathcal{O}(n^3)$ and the same complexity as the state-of-the-art methods \cite{16, 19, 26}.

We perform a comprehensive analysis of our approach, including the convergence proofs for the kernel-based graph regularized orthogonal multiplicative update rules. We conduct extensive experiments to compare our methods with other non-negative spectral clustering methods and further perform the sensitivity analysis of the parameters used in our approach. We highlight here the main contributions of the paper:

1. We formulate a nonlinear NMF with explicitly enforced orthogonality to address the subspace clustering problem.

2. We derive kernel-based orthogonal multiplicative updates to solve the constrained non-convex nonlinear NMF problem. We perform the convergence analysis for the multiplicative updates and give the convergence proofs using an auxiliary function approach \cite{Lee00algorithmsNMF}. 


3. We formulate a nonlinear (kernel-based) orthogonal graph regularized NMF approach to subspace clustering. The ability of the proposed method to exploit both the nonlinear nature of the manifold as well as its local geometric structure considerably improves the clustering performance.

4. The proposed clustering algorithms provide an insight into the connection between the spectral clustering methods and kernel NMF, where the kernel matrix in the kernel-based NMF multiplicative updates corresponds to the nonlinear graph affinity matrix in Ncut and Rcut spectral clustering.

The rest of the paper is organized as follows: in Section \ref{section1} we present a brief overview of the NMF-based spectral clustering. In Section \ref{section2}, we propose our framework and present three non-negative spectral clustering algorithms, along with the theoretical results on the equivalence of our approach and non-negative spectral clustering. In Section \ref{section3}, we compare our methods to the 9 recently proposed non-negative spectral clustering methods on 6 data sets. Lastly, we give the conclusions in Section \ref{section4}.

\section{Related work}
\label{section1}

We denote all matrices with bold upper case letters, all vectors with bold lower case letters. $\textbf{A}^\mathsf{T}$ denotes the transpose of the matrix $\textbf{A}$, and $\textbf{A}^{-1}$ denotes the inverse of the matrix $\textbf{A}$. $\textbf{I}$ denotes the identity matrix. The Frobenius norm is denoted as $\|\cdot\|_F$. The trace of the matrix is denoted with $\textrm{Tr}(\cdot)$. In Table 1 we summarize the rest of the notation.

%

\begin{table}[h!] \centering
	\renewcommand\thetable{1}
  \caption{\textit{Notations}} 
 \label{tab:data1}
  \scriptsize 
\begin{tabular}{@{\extracolsep{2pt}} ll} 
\hline 
\hline \\[-1.8ex] 
 Notation & Definition \\ \hline
    \textit{m} & the dimensionality of a data set \\
    \textit{n} & the number of data points \\ 
    \textit{k} & the number of clusters\\
    $\mathcal{L}$ & the Lagrangian\\
     $\textbf{K} \in \mathbb{R}^{n \times n}$ &the kernel matrix \\
    $\textbf{X} \in \mathbb{R}^{m \times n}$ & the input data matrix\\
   $\textbf{A} \in \mathbb{R}^{n \times n}$ & the graph affinity matrix\\
    $\textbf{D} \in \mathbb{R}^{n \times n}$ & the degree matrix based on $\textbf{A}$\\
    $\textbf{L} \in \mathbb{R}^{n \times n}$ & the graph Laplacian \\
    $\textbf{L}_{sym} \in \mathbb{R}^{n \times n}$ & the normalized graph Laplacian\\
       $\Phi(\textbf{X}) \in \mathbb{R}^{D \times n}$ & the nonlinear mapping\\
        $\textbf{H, Z} \in \mathbb{R}^{k \times n}$ & the cluster indicator matrices \\
        $\textbf{V} \in \mathbb{R}^{m \times k}$ & the basis matrix in input space\\
       $\textbf{F} \in \mathbb{R}^{n \times k}$ & the basis matrix in mapped space \\

\hline \\[-1.8ex] 
\end{tabular} 
\end{table} 
    
    \vspace{0.5cm}

\subsection{Definitions}

The task of subspace clustering is to find a low-dimensional subspace to fit each group of data points \cite{13, ChenpingHou2015, Ma2007, Rao2008}.  Let $\textbf{X} \in \mathbb{R}^{m \times n}$ denote the data matrix $m \times n$ which is comprised of $n$ data points $\textbf{x}_i \in \mathbb{R}^{m}$, drawn from a union of $k$ linear subspaces $S_1 \cup S_2 \cup ... \cup S_k$ of dimensions $\{m_i\}^k_{i=1}$.  Let $X_i \in \mathbb{R}^{m \times n_i}$ be a submatrix of $\textbf{X}$ of rank $m_i$ with $\sum_{i=1}^{k}n_i = n$. Given the input matrix $\textbf{X}$, subspace clustering assigns data points according to their subspaces. The first step is to construct a weighted similarity graph $G(V, E)$ from $\textbf{X}$, such that each node from the node set $V = \{1,2,...,n\}$ represents a data point $\textbf{x}_i  \in \mathbb{R}^{m}$ and each weighted edge represents a similarity based on distance (\textit{e.g}. the Euclidean distance) between the corresponding pair of nodes. Typical methods to construct the similarity graph are $\epsilon$-neighbourhood graphs, $k$-nearest neighbour graphs and fully connected graphs with Gaussian similarity function \cite{25,54}. Spectral clustering then finds the cluster membership of data points by using the spectrum of the graph Laplacian matrix. 
Let $\textbf{A} \in \mathbb{R}^{n \times n}$ be a symmetric affinity matrix of the graph and $A_{ij} \geq 0$ be the pairwise similarity between the nodes. The degree matrix $\textbf{D}$ based on $\textbf{A}$ is defined as the diagonal matrix with the degrees $d_1, ..., d_n$ on the diagonal, where the degree $d_i$ of a node $i$ is 
\begin{equation}
d_i = \sum_{j=1}A_{ij}
\end{equation}
Given a weighted graph $G(V, E)$ its unnormalized graph Laplacian matrix $\textbf{L}$ is given as $[45]$ 
\begin{equation}
\textbf{L} = \textbf{D} - \textbf{A}
\label{laplace}
\end{equation}
The symmetric normalized graph Laplacian matrix $\textbf{L}_{sym}$ is defined as
\begin{equation}
\textbf{L}_{sym} = \textbf{D}^{-1/2} \textbf{L} \textbf{D}^{-1/2}  = \textbf{I} - \textbf{D}^{-1/2} \textbf{A} \textbf{D}^{-1/2}
\label{laplace_sym}
\end{equation}
where  $\textbf{I}$ is the identity matrix. 

\subsection{Graph cuts}

The spectral clustering can be seen as partitioning a similarity graph  G$(V,E)$ into a set of nodes $S \subset V$ separated from the complementary set $\bar{S} = V\textbackslash S $. Depending on the choice of the function to optimize, the graph partition problem can be defined in different ways. The simplest choice of the function is the \textit{cut} $s(S,\bar{S})$ defined as $s(S,\bar{S}) = \sum_{v_i \in S, v_j \in \bar{S}}A_{ij}$. 
%
To achieve a better balance in the cardinality of $S$ and $\bar{S}$, the Ncut and Rcut optimization functions are proposed  \cite{33, 34, 35}.
Let $\textbf{h}_l$ be the indicator vector for cluster $C_l$, i.e. $\textbf{h}_l(i) = 1$ if $\textbf{x}_i \in C_l$, otherwise $\textbf{h}_l(i)=0,$ then $\textbar C_l \textbar =\textbf{h}_l \textbf{h}_l^\mathsf{T} .$ The cluster indicator matrix $\textbf{H} \in \mathbb{R}^{k \times n}$ can be defined as
\begin{equation}
\textbf{H}^\mathsf{T} = \left(\frac{\textbf{h}_1}{\|\textbf{h}_1\|},\frac{\textbf{h}_2}{\|\textbf{h}_2\|},...,\frac{\textbf{h}_k}{\|\textbf{h}_k\|}\right)
\end{equation}
Evidently, $\textbf{H}\textbf{H} ^\mathsf{T} = \textbf{I}$. 
Rcut spectral clustering can be formulated as the following optimization problem 
\begin{equation}
\underset{\textbf{H}}{\textrm{min}}\textrm{Tr}\left(\textbf{H}\textbf{L} \textbf{H}^\mathsf{T} \right) \hspace{5pt}s.t. \hspace{5pt} \textbf{H}\textbf{H}^\mathsf{T}=\textbf{I}
\end{equation}
where $\textrm{Tr}(\cdot)$ denotes the trace of a matrix and $\textbf{L}$ is the graph Laplacian. 
Similarly, define the cluster indicator vector as $\textbf{z}_k = \textbf{D}^{1/2} \textbf{h}_k  /\|\textbf{D}^{1/2} \textbf{h}_k \|$ and the cluster indicator
matrix as $\textbf{Z}^\mathsf{T} = \left(\textbf{z}_1,\textbf{z}_2,...,\textbf{z}_k\right)$ where $\textbf{Z} \in \mathbb{R}^{k \times n}$. Then Ncut is formulated as the minimization problem
\begin{equation}
\underset{\textbf{Z}}{\textrm{min} }\textrm{Tr}\left(\textbf{Z}\textbf{L}_{sym} \textbf{Z}^\mathsf{T} \right) \hspace{5pt}s.t. \hspace{5pt} \textbf{Z}\textbf{Z}^\mathsf{T}=\textbf{I}
\end{equation}
By allowing the cluster indicator matrices ($\textbf{H}$, $\textbf{Z}$) to be continuous valued the problem is solved by eigenvalue decomposition of the graph Laplacian matrix given in Eqs. (\ref{laplace}) and (\ref{laplace_sym}) \cite{3, 4,5}.

\subsection{NMF approach to non-negative spectral clustering}

The connection between the Ncut spectral clustering and symmetric NMF has been established in \cite{17} 
\begin{equation}
\textbf{D}^{-1/2} \textbf{A} \textbf{D}^{-1/2} = \textbf{H}^\mathsf{T} \textbf{H}, \hspace{5pt}s.t. \hspace{5pt}\textbf{H} \geq 0.
\label{ding008}
\end{equation}
According to the Theorem 2 from \cite{17}, enforcing symmetric factorization approximately retains the orthogonality of \textbf{H}. Similary, according to the Theorem 5 from \cite{18} the Rcut spectral clustering has been proved to be equivalent to the following symmetric NMF problem
\begin{equation}
\textbf{A} - \textbf{D} + \sigma \textbf{I} = \textbf{H}^\mathsf{T} \textbf{H}, \hspace{5pt}s.t.\hspace{5pt} \textbf{H}\textbf{H}^\mathsf{T} = \textbf{I}, \hspace{5pt}\textbf{H} \geq 0
\label{ding009}
\end{equation}
where $\sigma$ is the largest eigenvalue of the graph Laplacian matrix \textbf{L} and the matrix  $\textbf{H} \in \mathbb{R}^{k \times n}$ contains cluster membership information that data point
$\textbf{x}_i$ belongs to the cluster $c_i$  
\begin{equation}
c_i = \underset{1 \leq j \leq k}{\textrm{argmax}\hspace{3pt}\textbf{H}_{ji}}.
\end{equation}
In Eqs. (\ref{ding008}) and (\ref{ding009}) a factorization of $n \times n$ symmetric similarity matrix $\textbf{A}$ has a complexity $\mathcal{O}(kn^2)$ for $k$ clusters.

Based on the results \cite{17, 18}, in \cite{16} it is proved that for non-negative input data matrix $\textbf{X}$, and fully connected graph affinity matrix $\textbf{A}$ given as the standard inner product $\textbf{A}  = \textbf{X}^\mathsf{T}\textbf{X}$, Ncut spectral clustering is equivalent to the NMF of the scaled input data matrix (NSC-Ncut)
\begin{equation}
\textbf{D}^{-1/2}\textbf{X}^\mathsf{T} \approx \textbf{Z}^\mathsf{T} \textbf{Y} \hspace{5pt}s.t. \hspace{5pt} \textbf{Z} \textbf{Z}^\mathsf{T}= \textbf{I},  \textbf{Z} \geq 0
\label{ncutnsc}
\end{equation}
with cluster indicator matrix $\textbf{Z} \in \mathbb{R}^{k \times n}$. Similarly, the Theorem 2 \cite{16} the establishes the connection of Rcut non-negative spectral clustering (NSC-Rcut) and NMF problem 
\begin{equation}
\textbf{X}^\mathsf{T} \approx \textbf{H}^\mathsf{T} \textbf{Y} \hspace{5pt}s.t. \hspace{5pt} \textbf{H}\textbf{H}^\mathsf{T}= \textbf{I},  \textbf{H} \geq 0
\label{rcutnsc}
\end{equation}
with cluster indicator matrix $\textbf{H}  \in \mathbb{R}^{k \times n}$. Both NMF-based approaches to non-negative spectral clustering  (\ref{ncutnsc}) and (\ref{rcutnsc}) are formulated in the input data space as a factorization of an input data matrix  $\textbf{X} \in \mathbb{R}^{m \times n}$ with the complexity $\mathcal{O}(nmk)$ \cite{16}. The matrix factorization in Eqs. (\ref{ncutnsc}) and (\ref{rcutnsc}) is limited to the graph affinity matrix defined as an inner product of the input data matrix.


Furthermore, the global discriminative NMF-based NSC model introduced in \cite{19}, includes an additional nonlinear discriminative regularization term to the NMF optimization function proposed in \cite{16}. As shown in \cite{19}, the global discriminant information greatly improves the accuracy of NSC-Ncut and NSC-Rcut \cite{16}. Although in \cite{19} the nonlinear character of the manifold is taken into account through the nonlinear discriminative matrix, the NMF data fidelity terms are still defined in the input data space.

\section{Nonlinear orthogonal NMF approach to subspace clustering}
\label{section2}

In this section we develop a nonlinear orthogonal NMF approach to subspace clustering and establish its equivalence with Ncut and Rcut spectral clustering algorithms. We generalize the NMF objective function to a nonlinear transformation of the input data and derive kernel-based NMF update rules with explicitly imposed orthogonality constraints on the clustering matrix  $\textbf{H}$ (or $\textbf{Z}$). Enforcing the explicit orthogonality into the multiplicative rules allows obtaining the cluster membership directly from the cluster indicator matrix. In this way, we obtain a formulation of the nonlinear NMF that explicitly addresses the subspace clustering problem.

\subsection{Kernel-based orthogonal NMF mutiplicative updates}

In this paper we emphasize the orthogonality of the nonlinear NMF to keep the clustering interpretation while taking into account the nonlinearity of the space data are drawn from. We enforce rigorous orthogonality constraint into the NMF optimization problem and seek to obtain kernel-based orthogonal multiplicative update rules to solve it. 

Let $\textbf{X} = (\textbf{x}_1, \textbf{x}_2, ... \textbf{x}_n) \in \mathbb{R}^{m \times n}$ be the data matrix of non-negative elements. The NMF factorizes \textbf{X} into two low-rank non-negative matrices
\begin{equation}
\textbf{X} \approx \textbf{VH}
\end{equation}
where $\textbf{V} = (\textbf{v}_1, \textbf{v}_2,...,\textbf{v}_k) \in \mathbb{R}^{m \times k}$ and $\textbf{H}^\mathsf{T} = (\textbf{h}_1, \textbf{h}_2,...,\textbf{h}_k) \in \mathbb{R}^{n \times k}$  and $k$ is a prespecified rank parameter. Generally, the rank of matrices \textbf{V} and \textbf{H} is much lower than the rank of \textbf{X} (i.e., $k \ll \textrm{min}(m, n))$. The non-negative matrices \textbf{V} and \textbf{H} are obtained by solving the following minimization problem
\begin{equation}
\underset{\textbf{V}, \textbf{H} \geq 0}{\textrm{min}} \|\textbf{X} - \textbf{VH}\|^2_F 
\label{nmf}
\end{equation}
Consider now a nonlinear transformation (a mapping) to the higher $D$-dimensional (or infinite) space $\textbf{x}_i \rightarrow \Phi(\textbf{x}_i)$ or  $\textbf{X} \rightarrow \Phi(\textbf{X}) = (\Phi(\textbf{x}_1), \Phi(\textbf{x}_2), ... , \Phi(\textbf{x}_n))$ $\in \mathbb{R}^{D \times n}$. The nonlinear NMF problem aims to find two non-negative matrices $\textbf{W}$ and $\textbf{H}$ whose product can approximate the mapping of the original matrix $\Phi(\textbf{X})$ 
\begin{equation}
\Phi(\textbf{X}) \approx \textbf{W} \textbf{H}
\label{nmfphi}
\end{equation}
For instance, we can consider  nonlinear data set composed of two rings as in Fig. 1. The standard linear NMF (\ref{nmf}) \cite{22} is not able to separate the two nonlinear clusters. Compared with the solution of Eq.  (\ref{nonlinmin}), the nonlinear NMF is able to produce the nonlinear separating hypersurfaces between the clusters. We formulate the objective function for the nonlinear orthogonal NMF as
\begin{figure*}
\center
\includegraphics[width=0.7\textwidth]{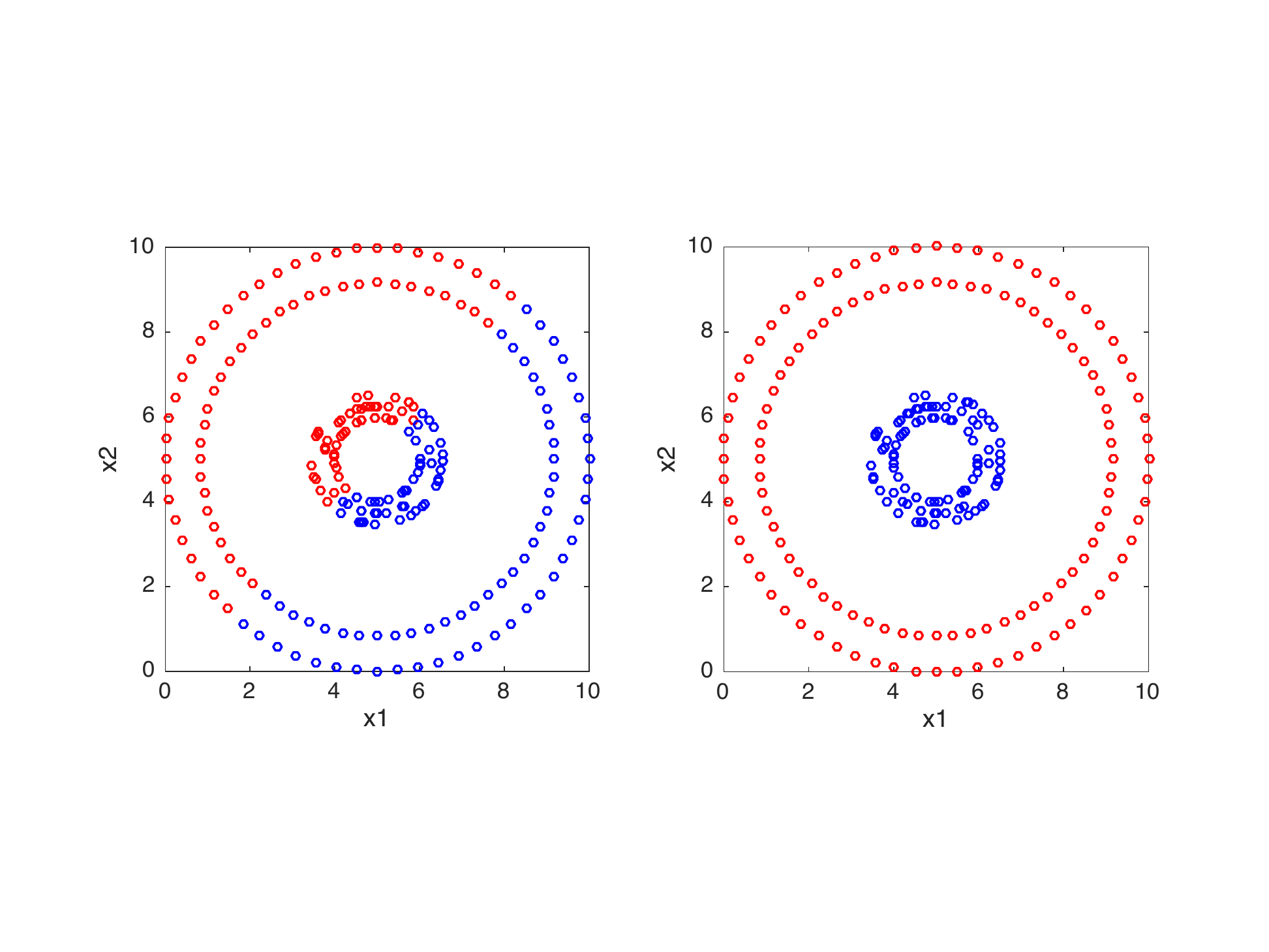}
\caption{Clustering with NMF (left) and nonlinear NMF (right). We apply the nonlinear NMF (KNSC-Ncut) (35) with Gaussian kernel (right) and linear NMF introduced in \cite{20} to the synthetic data set composed of two rings and denote the cluster memberships with different colors. The nonlinear NMF is able to produce the nonlinear separating hypersurfaces between the two rings.}
\label{AAA}
\end{figure*}
\begin{equation}
\underset{\textbf{H,F} \geq 0}{\textrm{min}} \|\Phi(\textbf{X}) - \textbf{W}\textbf{H}\|^2_F\hspace{5pt}s.t. \hspace{5pt}\textbf{H}\textbf{H}^\mathsf{T}= \textbf{I} 
\label{nonlinmin}
\end{equation}

Here, $\textbf{W}$ is the basis in feature space and $\textbf{H}$ is the clustering matrix.  It is worth noting that since $\Phi$ can be infinite dimensional, it is impossible to directly factorize $\Phi(\textbf{X})$ \cite{48, 49, 50}.  In what follows we will derive a practical method to solve this problem, and keep the rigorous orthogonality imposed on the clustering matrix. Following \cite{50} we restrict  $\textbf{W}$ to be a linear combination of transformed input data points, i.e., assume that $ \textbf{W}$ lies in the column space of $\Phi(\textbf{X})$
\begin{equation}
 \textbf{W} = \Phi(\textbf{X})\textbf{F}
\label{trb}
\end{equation}
The equation (\ref{trb}) can be interpreted as a simple transformation to the new basis, leading to the following minimization problem 
\begin{equation}
\underset{\textbf{H,F} \geq 0}{\textrm{min}} \|\Phi(\textbf{X}) - \Phi(\textbf{X})\textbf{F}\textbf{H}\|^2_F,\hspace{5pt}s.t. \hspace{5pt}\textbf{H}\textbf{H}^\mathsf{T}= \textbf{I}
\label{nonlinmin}
\end{equation}
The optimization problem of Eq.  (\ref{nonlinmin}) is convex in either $\textbf{F}$ or $\textbf{H}$, but not in both, meaning that the algorithm can only guarantee convergence to a local minimum \cite{123}. The standard way to optimize (\ref{nonlinmin}) is to adopt an iterative, two-step strategy to alternatively optimize $(\textbf{F},\textbf{H})$.  At each iteration, one of the matrices $(\textbf{F},\textbf{H})$ is optimized while the other one is fixed. The resulting multiplicative update rules with explicitly included orthogonality constraints are obtained as

\begin{equation}
\label{eq:Hupdate}
H_{ij} \leftarrow H_{ij} \frac{(\alpha \textbf{F}^\mathsf{T}\textbf{K} + 2\mu \textbf{H})_{ij}}{(\alpha \textbf{F}^\mathsf{T} \mathbf{K} \textbf{FH }+ 2\mu  \textbf{HH}^\mathsf{T} \textbf{H})_{ij}}
\end{equation}
\begin{equation}
\label{eq:Fupdate}
F_{jl} \leftarrow F_{jl} \frac{(\mathbf{K} \textbf{H}^\mathsf{T})_{jl}}{(\mathbf{K} \textbf{FHH}^\mathsf{T})_{jl}}
\end{equation}
where $\textbf{K} \in \mathbb{R}^{n \times n}$ is the kernel matrix \cite{ScholkopfKernel, KoprivaKernel} defined as $\textbf{K} \equiv \Phi^\mathsf{T}(\textbf{X})\Phi(\textbf{X})$, where $\Phi(\textbf{X})$ is a feature matrix in a nonlinear infinite feature space. 

We discuss two issues: (\textit{i}) convergence of the algorithm, (\textit{ii}) correctness of the converged solution. \textbf{Correctness.} The correctness of the solution is assured by the fact that the solution at convergence will satisfy the Karush-Kahn-Tucker (KKT) conditions for (\ref{nonlinmin}). The Lagrangian $\mathcal{L}$ of the the above optimization problem (\ref{nonlinmin}) is 
\begin{equation}
 \mathcal{L} = \alpha \textrm{Tr}[\Phi(\textbf{X})\Phi^\mathsf{T}(\textbf{X})] - 2\alpha \textrm{Tr}[\Phi(\textbf{X})\textbf{FH}\Phi^\mathsf{T}(\textbf{X})]  + \alpha \textrm{Tr}[\Phi(\textbf{X})\textbf{FHH}^\mathsf{T}\textbf{F}^\mathsf{T}\Phi^\mathsf{T}(\textbf{X})] +  \mu \|\textbf{HH}^\mathsf{T} - \textbf{I}_k \|^2_F
 \label{rew}
\end{equation}
By computing the partial derivatives of (\ref{rew})  with respect to $\textbf{H}$ and $\textbf{F}$, we obtain
\begin{equation}
\frac{\partial \mathcal{L}}{\partial \textbf{H}} = -2\alpha \textbf{F}^\mathsf{T}\Phi^\mathsf{T}(\textbf{X})\Phi(\textbf{X})+ 2\alpha \textbf{F}^\mathsf{T} \Phi^\mathsf{T}(\textbf{X})\Phi(\textbf{X}) \textbf{FH} + 4 \mu \textbf{H}(\textbf{H}^\mathsf{T}\textbf{H} - \textbf{I}_{n \times n})
\label{partial}
\end{equation}
\begin{equation}
\frac{\partial \mathcal{L}}{\partial \textbf{F}} = - \alpha\Phi^\mathsf{T}(\textbf{X})\Phi(\textbf{X}) \textbf{H}^\mathsf{T}+\alpha \Phi^\mathsf{T}(\textbf{X})\Phi(\textbf{X}) \textbf{FHH}^\mathsf{T} 
\label{urF}
\end{equation}
Substituting the quadratic terms with the kernel matrix $\textbf{K} = \Phi^\mathsf{T}(\textbf{X})\Phi(\textbf{X})$ yields 
\begin{equation}
 \alpha (\textbf{F}^\mathsf{T}\textbf{K} \textbf{FH} -\textbf{F}^\mathsf{T} \textbf{K}) + 2\mu \textbf{H}(\textbf{H}^\mathsf{T}\textbf{H} - \textbf{I}_{n \times n}) = 0
\end{equation}
\begin{equation}
 -2\alpha\textbf{K} \textbf{H}^\mathsf{T}+ 2\alpha \textbf{K} \textbf{FHH}^\mathsf{T}= 0
\end{equation}
Defining the Lagrange multiplier matrix for constraint $\textbf{H} \geq 0$ as $\Psi = [\psi_{ij}]$ gives the KKT condition $\psi_{ij} H_{ij} = 0$.  Similarly, the Lagrange multiplier matrix for constraint $\textbf{F} \geq 0$ is given by $\Xi = [\xi_{jl}]$ and  $\xi_{ij} F_{ij} = 0$. We obtain
\begin{equation}
 [\alpha (\textbf{F}^\mathsf{T}\textbf{K} \textbf{FH} -\textbf{F}^\mathsf{T} \textbf{K}) + 2\mu \textbf{H}(\textbf{H}^\mathsf{T}\textbf{H} - \textbf{I}_{n \times n})]_{ij} H_{ij} = 0
\end{equation}
\begin{equation}
 [2\alpha \textbf{K} \textbf{FHH}^\mathsf{T} - 2\alpha \textbf{K} \textbf{H}^\mathsf{T}]_{jl} F_{jl} = 0
\end{equation}

Separating positive and negative parts of the gradient leads to the multiplicative update rules (\ref{eq:Fupdate}) and (\ref{eq:Hupdate}). 


\textbf{Convergence.} The convergence is proved by following the  auxiliary function method in  \cite{50,18}.
As shown in \cite{50}, these update rules guarantee the decrease of the error and eventual convergence to local minima. Note that in \cite{50} a more general proof of the convergence can be obtained, for semi-nonnegative matrix factorization, where input data matrix is negative $\textbf{X} \textless 0$.
We provide the proof for the convergence in the Appendix B.

\subsection{Kernel-based orthogonal NMF and spectral clustering}


 A connection between spectral clustering and factorization of the graph affinity matrix $\textbf{A}$ was demonstrated in \cite{17} for Ncut spectral clustering, and for Rcut spectral clustering in \cite{18}. It was also shown that the spectral clustering can be viewed as a factorization of the (scaled) data matrix itself \cite{16}. Our question is whether the spectral clustering can be viewed as a non-negative factorization of the input data matrix mapped to a nonlinear feature space. From Eq. (12) it can be seen that the Ncut spectral clustering is equivalent to the optimization problem
\begin{equation}
\underset{\textbf{Z} \geq 0}{\textrm{max}}\textrm{ Tr}\left(\textbf{Z}\textbf{D}^{-1/2}\textbf{A} \textbf{D}^{-1/2}\textbf{Z}^\mathsf{T} \right) \hspace{5pt}s.t. \hspace{5pt}   \textbf{Z}\textbf{Z}^\mathsf{T}=\textbf{I}
\label{ncuteq}
\end{equation}
\textbf{Theorem 1.} \textit{Let }$\textbf{X} \geq 0$ \textit{denote the input data matrix. Let the similarity between the data points be defined as the inner product in the nonlinear feature space, i.e. the graph affinity matrix} $\textbf{A} = \Phi^\mathsf{T}(\textbf{X})\Phi(\textbf{X})$. \textit{Then the }$k$\textit{-way Ncut spectral clustering (\ref{ncuteq}) is equivalent to the non-negative matrix factorization of the scaled input data matrix mapped to the nonlinear feature space} $\Phi(\textbf{X}) \textbf{D}^{-1/2} = \textbf{WZ}$ \textit{subject to} $\textbf{Z}  \textbf{Z}^\mathsf{T} = \textbf{I}$, where
$\textbf{W} = \Phi(\textbf{X})\textbf{F}$ and $\textbf{Z}$ and $\textbf{F}$ \textit{are two non-negative matrices, and the columns of} $\textbf{Z}$ \textit{serve as a clustering indicator vector of each data point.}

The proof of the Theorem 1 is given in the Appendix A. Theorem 1 shows that Ncut spectral clustering can be viewed as a nonlinear orthogonal NMF problem with the scaling factor $\textbf{D}^{-1/2}$. For the Rcut spectral clustering we cannot obtain an exact equivalence. However, we can relax the Rcut spectral clustering and get an equivalence between the relaxed Rcut spectral clustering and nonlinear orthonormal NMF.

\textbf{Theorem 2.}\textit{ Let} $\textbf{X} \geq 0$ \textit{denote the input data matrix. Let the similarity between the data points be defined by inner product in nonlinear feature space i.e.  the affinity matrix }$\textbf{A} = \Phi^\mathsf{T}(\textbf{X})\Phi(\textbf{X})$. \textit{Then the} $k$-\textit{way relaxed Rcut spectral clustering (\ref{rcutnsc}) is equivalent to the non-negative matrix factorization of the data matrix }$\Phi(\textbf{X}) = \textbf{WH}$ \textit{subject to} $\textbf{H} \textbf{H}^\mathsf{T} = \textbf{I}$, \textit{where}
$\textbf{W} = \Phi(\textbf{X})\textbf{F}$ and $\textbf{H}$ and $\textbf{F}$ \textit{are two non-negative matrices, and the columns of $\textbf{H}$ serve as a clustering indicator vector of each data point. }

The proof of the Theorem 2 is given in the Appendix A. Theorems 1 and 2 establish the nonlinear orthogonal NMF approach to non-negative spectral clustering. Our assumptions include that the similarity graph is fully connected and the similarity matrix $\textbf{A}$ is given by the kernel $\textbf{K} = \Phi^\mathsf{T}(\textbf{X})\Phi(\textbf{X})$. Similarly to this result, it was shown  in \cite{17} that the standard inner-product matrix $\textbf{A} = \textbf{X}^\mathsf{T}\textbf{X}$ can be extended to any other kernel by a nonlinear transformation to a higher dimensional space. 

To solve Ncut and Rcut spectral clustering we employ the kernel-based multiplicative update rules with orthonormal constraints. Considering the equivalence and solving the two optimization problems we obtain \textit{kernel-based non-negative spectral clustering} for Ncut (KNSC-Ncut)
 \begin{equation}
\underset{\textbf{Z,F} \geq 0}{\textrm{min}} \|\Phi(\textbf{X})\textbf{D}^{-1/2} - \Phi(\textbf{X})\textbf{F}\textbf{Z}\|^2_F,\hspace{5pt}s.t. \hspace{5pt}\textbf{Z}\textbf{Z}^\mathsf{T}= \textbf{I}
 \end{equation}
 with the following multiplicative update rule
 
\begin{equation}
\label{eq:HupdateNcut}
Z_{ij} \leftarrow Z_{ij} \frac{(\alpha \textbf{F}^\mathsf{T}\textbf{K}\textbf{D}^{-1/2} + 2\mu \textbf{Z} )_{ij}}{(\alpha \textbf{F}^\mathsf{T} \textbf{K} \textbf{FZ}+ 2\mu  \textbf{ZZ}^\mathsf{T} \textbf{Z} )_{ij}}
\end{equation}

\begin{equation}
\label{eq:FupdateNcut}
F_{jl} \leftarrow F_{jl} \frac{(\textbf{K} \textbf{Z}^\mathsf{T})_{jl}}{(\textbf{K} \textbf{FZZ}^\mathsf{T})_{jl}}
\end{equation}

The parameter $\mu$ can be set so that the orthogonality of the matrix $\textbf{Z}$ is preserved during the updates. An exact orthogonality of the clustering matrix \textbf{Z} implies each column of \textbf{Z} can have only one non-zero element, which implies that each data object belongs only to one cluster.  This is hard clustering, such as in $k$-means  \cite{17,31}. Furthermore, KNSC-Ncut has a soft clustering intepretation \cite{20, 18, 17} where a data point could belong fractionally to more than one cluster. The soft clustering membership of data point $\textbf{x}_i$ to cluster $j$ can be defined as a probability distribution $c_{i,j} = \textbf{Z}_{ji} / \sum_k \textbf{Z}_{ki}$. We summarize the KNSC-Ncut algorithm in the Algorithm 1. Similarly, the optimization problem for \textit{kernel-based non-negative spectral clustering} for Rcut (KNSC-Rcut)
 \begin{equation}
\underset{\textbf{H,F} \geq 0}{\textrm{min}} \|\Phi(\textbf{X}) - \Phi(\textbf{X})\textbf{F}\textbf{H}\|^2_F,\hspace{5pt}s.t. \hspace{5pt}\textbf{H}\textbf{H}^\mathsf{T}= \textbf{I}
 \end{equation}
gives the multiplicative update rule for KNSC-Rcut
\begin{equation}
\label{eq:Hupdate}
H_{ij} \leftarrow H_{ij} \frac{(\alpha \textbf{F}^\mathsf{T}\textbf{K} + 2\mu \textbf{H})_{ij}}{(\alpha \textbf{F}^\mathsf{T} \textbf{K} \textbf{FH }+ 2\mu  \textbf{HH}^\mathsf{T} \textbf{H} )_{ij}}
\end{equation}
\begin{equation}
\label{eq:Fupdate}
F_{jl} \leftarrow F_{jl} \frac{(\textbf{K} \textbf{H}^\mathsf{T})_{jl}}{(\textbf{K} \textbf{FHH}^\mathsf{T})_{jl}}
\end{equation}
and summarize the KNSC-Rcut algorithm in Algorithm 2.
 {
\begin{algorithm}
\captionof{algorithm}{Kernel-based non-negative spectral clustering for Ncut (KNSC-Ncut)} 
\begin{algorithmic}
\State \textbf{Input:} \textbf{X} $\in  \mathbb{R}^{m \times n}$, $\textbf{K} \in  \mathbb{R}^{n \times n}$,  $\textbf{A} \in  \mathbb{R}^{n \times n}$, number of clusters \textit{k}
\State \textbf{Output:} clustering matrix \textbf{Z} $\in  \mathbb{R}^{k \times n}$, vector of cluster memberships $c_i = \underset{1 \leq j \leq k}{\textrm{argmax}\hspace{3pt}\textbf{Z}_{ji}}$\\

Initialize two non-negative matrices $\textbf{Z} \in \mathbb{R}^{k \times n}$  and $\textbf{F} \in \mathbb{R}^{n \times k}$ with random numbers generated in the range $[0,1]$. \\
\State Calculate the degree matrix $\textbf{D} = \textrm{diag}(d_1,...d_n)$ 
\begin{equation}
d_i = \sum_{j=1}A_{ij}
\end{equation}
\Repeat 
\State
\begin{equation*}
Z_{ij} \leftarrow Z_{ij}\frac{(\alpha \textbf{F}^\mathsf{T}\textbf{K}\textbf{D}^{-1/2} + 2\mu \textbf{Z} )_{ij}}{(\alpha \textbf{F}^\mathsf{T} \textbf{K} \textbf{FZ}+ 2\mu  \textbf{ZZ}^\mathsf{T} \textbf{Z} )_{ij} }
\end{equation*}
\State 
\begin{equation*}
F_{jl} \leftarrow F_{jl} \frac{(\textbf{K} \textbf{Z}^\mathsf{T})_{jl}}{(\textbf{K} \textbf{FZZ}^\mathsf{T})_{jl}}
\end{equation*}

\Until{Stopping criterion is reached}
\end{algorithmic}
\end{algorithm}}

{\begin{algorithm}
\captionof{algorithm}{Kernel-based non-negative spectral clustering for Rcut (KNSC-Rcut)} 
\begin{algorithmic}
\State \textbf{Input:} \textbf{X} $\in  \mathbb{R}^{m \times n}$, $\textbf{K} \in  \mathbb{R}^{n \times n}$, number of clusters \textit{k}
\State \textbf{Output:} clustering matrix $\textbf{H}$ $\in  \mathbb{R}^{k \times n}$, vector of cluster memberships $c_i = \underset{1 \leq j \leq k}{\textrm{argmax}\hspace{3pt}\textbf{H}_{ji}}$\\

Initialize two non-negative matrices $\textbf{H} \in \mathbb{R}^{k \times n}$  and $\textbf{F} \in \mathbb{R}^{n \times k}$ with random numbers generated in the range $[0,1]$. 
\Repeat 
\begin{equation*}
H_{ij}\leftarrow H_{ij} \frac{(\alpha \textbf{F}^\mathsf{T}\textbf{K} + 2\mu \textbf{H})_{ij}}{(\alpha \textbf{F}^\mathsf{T} \textbf{K} \textbf{FH }+ 2\mu  \textbf{HH}^\mathsf{T} \textbf{H} )_{ij}}
\end{equation*}
\State 
\begin{equation*}
F_{jl} \leftarrow F_{jl}\frac{(\textbf{K} \textbf{H}^\mathsf{T})_{jl}}{(\textbf{K} \textbf{FHH}^\mathsf{T})_{jl}}
\end{equation*}
\Until{Stopping criterion is reached}
\end{algorithmic}
\end{algorithm}
}
 The convergence of the multiplicative update rules (\ref{eq:HupdateNcut})-- (\ref{eq:FupdateNcut}), and (\ref{eq:Hupdate})--(\ref{eq:Fupdate}), has been proved in Appendix B by the auxiliary function method. These update rules guarantee the decrease of error and eventually converge to a local minima \cite{50}. In our experiments, we have set the maximum amount of iterations to 300 (usually 100 iterations are enough) and we use the convergence rule $E_{i-1} - E_i \leq \kappa \hspace{2pt} \textrm{max}(1,E_{i-1}) $  in order to stop the updates when the reconstruction error ($E_i$) between the current and previous update is small enough. We have set the $\kappa = 10^{-3}$. 

The two proposed algorithms have a run-time complexity of $\mathcal{O}(kn^2)$ for clustering $n$ data points to $k$ clusters, which is less than standard spectral clustering methods $\mathcal{O}(n^3)$ and the same complexity as the state-of-the-art methods \cite{16, 19, 26}. The main advantage of the kernel-based NMF approach is that it can be easily optimized to achieve higher clustering accuracy for the data drawn from nonlinear manifolds, avoiding the computation of eigenvalues and eigenvectors.



\subsection{Graph regularized kernel-based orthogonal NMF}
\label{grreg}

A non-negative matrix factorization that respects the geometric structure of the data in the nonlinear feature space can be constructed by introducing an additional graph regularization term into the objective function (\ref{nonlinmin}).
Recall that our nonlinear NMF tries to find a set of basis vectors that can be used to best approximate the data $\Phi(\textbf{X}) = \textbf{W} \textbf{H}$. Let $\textbf{h}_j$ denote the $j$-th column of $\textbf{H}$, $\textbf{h}_j = [h_{j1},...,h_{jk}]$, then $\textbf{h}_j$  can be regarded as the new representation of the $j$-th data point with respect to the new basis $\textbf{W} =  \Phi(\textbf{X})\textbf{F}$. The graph regularization term can be viewed as a local invariance assumption \cite{54, 7p, 19p}, which states that if two data points $\Phi(\textbf{x}_i)$  and $\Phi(\textbf{x}_j)$ are close to each other in the original geometry of the data distribution, then  $\textbf{h}_j$ and $\textbf{h}_l$, the low dimensional representations of this two points, are also close to each other. This can be written as
\begin{equation}
\mathcal{R} = \frac{1}{2} \sum_{j,l = 1}^{n} \|\textbf{h}_j - \textbf{h}_l \|_{F}^2 \mathcal{A}_{jl} = \sum_{j=1}^n \textbf{h}_j \textbf{h}_j^\mathsf{T} \textbf{D}_{j,j} - \sum_{j,l=1}^n \textbf{h}_j \textbf{h}_l^\mathsf{T} \textbf{A}_{j,l} = \textrm{Tr}(\textbf{H} \textbf{L} \textbf{H}^\mathsf{T})
\label{graf}
\end{equation}
By minimizing the regularization term $\mathcal{R}$ with respect to $\textbf{H}$,  we expect that when $\Phi(\textbf{x}_i)$  and $\Phi(\textbf{x}_j)$ are close (\textit{i.e.} when $\textbf{A}_{jl}$ is large) the points $\textbf{h}_j$ and $\textbf{h}_l$ are also close with respect to the new basis. 
The objective function for nonlinear orthogonal graph regularized NMF is given as
\begin{equation}
\underset{\textbf{H,F} \geq 0}{\textrm{min}} \alpha \|\Phi(\textbf{X}) - \Phi(\textbf{X})\textbf{F}\textbf{H}\|^2_F +  \lambda \textrm{Tr}(\textbf{H} \textbf{L} \textbf{H}^\mathsf{T}) ,\hspace{2pt}s.t. \hspace{3pt}\textbf{H}\textbf{H}^\mathsf{T}= \textbf{I}
\label{nonlinmingr}
\end{equation}
By adopting the same iterative procedure to alternatively fix one of the matrices \textbf{F} and \textbf{H}, we solve the minimization problem (\ref{nonlinmingr}) and obtain the multiplicative update rules 
\begin{equation}
\label{eq:rulesgraph1}
H_{ij} \leftarrow H_{ij} \frac{(\alpha \textbf{F}^\mathsf{T}\textbf{K}+ 2\mu \textbf{H} + \lambda \textbf{H}\textbf{A})_{ij}}{(\alpha \textbf{F}^\mathsf{T} \textbf{K} \textbf{FH}+ 2\mu  \textbf{HH}^\mathsf{T} \textbf{H} + \lambda \textbf{H}\textbf{D})_{ij}}
\end{equation}
\begin{equation}
\label{eq:rulesgraph2}
F_{jl} \leftarrow F_{jl} \frac{(\textbf{K} \textbf{H}^\mathsf{T})_{jl}}{(\textbf{K} \textbf{FHH}^\mathsf{T})_{jl}}
\end{equation}
where $\textbf{K}$ is the kernel matrix. There are many choices to define the weight matrix $\textbf{A}$ of the graph. 
For example, the scalar product weighting and the cosine similarity are most suitable for processing documents, while for image data the heat kernel is commonly used \cite{23, 25, 54}. 
We will use the fully connected affinity graph with the Gauss kernel weighting, as we do not treat different weighting schemes separately. 

\textbf{Correctness.} The correctness of the solution is assured by the fact that the solution at convergence will satisfy the KKT conditions for the optimization problem (\ref{nonlinmingr}). The Lagrangian $\mathcal{L}$ of the the optimization problem (\ref{nonlinmingr}) can be written as
\begin{equation*}
 \mathcal{L} = \alpha \textrm{Tr}[\Phi(\textbf{X})\Phi^\mathsf{T}(\textbf{X})] - 2\alpha \textrm{Tr}[\Phi(\textbf{X})\textbf{FH}\Phi^\mathsf{T}(\textbf{X})]  + \alpha \textrm{Tr}[\Phi(\textbf{X})\textbf{FHH}^\mathsf{T}\textbf{F}^\mathsf{T}\Phi^\mathsf{T}(\textbf{X})] +
\end{equation*}
\begin{equation}
 +  \mu \|\textbf{HH}^\mathsf{T} - \textbf{I}_k \|^2_F +  \lambda \textrm{Tr}[\textbf{H} \textbf{D}\textbf{H}^\mathsf{T}] - \lambda \textrm{Tr}[\textbf{H} \textbf{A}\textbf{H}^\mathsf{T}]
 \label{lagrgr}
\end{equation}

We calculate the partial derivatives of  (\ref{lagrgr}) with respect to $\textbf{H}$ and $\textbf{F}$ 
\begin{equation}
\frac{\partial \mathcal{L}}{\partial \textbf{H}} = -2\alpha \textbf{F}^\mathsf{T}\Phi^\mathsf{T}(\textbf{X})\Phi(\textbf{X})+ 2\alpha \textbf{F}^\mathsf{T} \Phi^\mathsf{T}(\textbf{X})\Phi(\textbf{X}) \textbf{FH} + 4 \mu \textbf{H}(\textbf{H}^\mathsf{T}\textbf{H} - \textbf{I}_{n \times n}) + 2\lambda \textbf{H} \textbf{D}- 2\lambda \textbf{H} \textbf{A}
\label{partialgr}
\end{equation}
\begin{equation}
\frac{\partial \mathcal{L}}{\partial \textbf{F}} = - \alpha\Phi^\mathsf{T}(\textbf{X})\Phi(\textbf{X}) \textbf{H}^\mathsf{T}+\alpha \Phi^\mathsf{T}(\textbf{X})\Phi(\textbf{X}) \textbf{FHH}^\mathsf{T} 
\label{urFgr}
\end{equation}
Substituting the quadratic terms with kernel matrix gives 
\begin{equation}
 \alpha (\textbf{F}^\mathsf{T}\textbf{K} \textbf{FH} -\textbf{F}^\mathsf{T} \textbf{K}) + 2\mu \textbf{H}(\textbf{H}^\mathsf{T}\textbf{H} - \textbf{I}_{n \times n}) + \lambda\textbf{HL}= 0
\end{equation}
\begin{equation}
 -2\alpha\textbf{K} \textbf{H}^\mathsf{T}+ 2\alpha \textbf{K} \textbf{FHH}^\mathsf{T}= 0
\end{equation}
Defining the Lagrange multiplier matrix for constraint $\textbf{H} \geq 0$ as $\Psi = [\psi_{ij}]$, the KKT condition is $\psi_{ij} H_{ij} = 0$.  Similarly, the Lagrange multiplier matrix for constraint $\textbf{F} \geq 0$ is given by $\Xi = [\xi_{jl}]$ and we obtain 
\begin{equation*}
 [\alpha (\textbf{F}^\mathsf{T}\textbf{K} \textbf{FH} -\textbf{F}^\mathsf{T} \textbf{K}) + 2\mu \textbf{H}(\textbf{H}^\mathsf{T}\textbf{H} - \textbf{I}_{n \times n}) + \lambda \textbf{HL}]_{ij} H_{ij} = 0
\end{equation*}
\begin{equation}
 [2\alpha \textbf{K} \textbf{FHH}^\mathsf{T} - 2\alpha \textbf{K} \textbf{H}^\mathsf{T}]_{jl} F_{jl} = 0
\end{equation}
We separate positive and negative parts of the gradient and obtain multiplicative update rules (\ref{eq:rulesgraph1}) and (\ref{eq:rulesgraph2}). 
By setting $\lambda = 0$ the update rules in Eq. (\ref{eq:rulesgraph1}) and (\ref{eq:rulesgraph2}) reduce to the update rules of the KONMF. We summarize the graph regularized kernel-based orthogonal NMF in the Algorithm 3.

{
\begin{algorithm}
\captionof{algorithm}{\textbf{Kernel-based orthogonal graph regularized NMF (KOGNMF)}} 
\begin{algorithmic}
\State \textbf{Input:} \textbf{X} $\in  \mathbb{R}^{m \times n}$, number of clusters \textit{k}, $\textbf{K} \in  \mathbb{R}^{n \times n}$, $\textbf{A} \in  \mathbb{R}^{n \times n}$
\State \textbf{Output:} clustering matrix $\textbf{H}$, vector of cluster memberships $c_i = \underset{1 \leq j \leq k}{\textrm{argmax}\hspace{3pt}\textbf{H}_{ji}}$\\

Initialize two non-negative matrices $\textbf{H} \in \mathbb{R}^{k \times n}$  and $\textbf{F} \in \mathbb{R}^{n \times k}$ with random numbers generated in the range $[0,1]$. \\

\State Calculate the degree matrix $\textbf{D} = \textrm{diag}(d_1,...d_n)$ 
\begin{equation*}
d_i = \sum_{j=1}A_{ij}
\end{equation*}
\Repeat 
\begin{equation*}
H_{ij} \leftarrow H_{ij} \frac{(\alpha \textbf{F}^\mathsf{T}\textbf{K}+ 2\mu \textbf{H} + \lambda \textbf{H}\textbf{A})_{ij}}{(\alpha \textbf{F}^\mathsf{T} \textbf{K} \textbf{FH}+ 2\mu  \textbf{HH}^\mathsf{T} \textbf{H} + \lambda \textbf{H}\textbf{D})_{ij} } 
\end{equation*}
\State
\begin{equation*}
F_{jl} \leftarrow F_{jl} \frac{(\textbf{K} \textbf{H}^\mathsf{T})_{jl}}{(\textbf{K} \textbf{FHH}^\mathsf{T})_{jl}}
\end{equation*}
\Until{Stopping criterion is reached}
\end{algorithmic}
\end{algorithm}

}

The proposed algorithm has two additional matrix multiplications $\textbf{H}\textbf{A}$ and $\textbf{H}\textbf{D}$ with complexity of $O(kn^2)$. Therefore, the total run-time complexity is unchanged and equal to $O(kn^2)$ for clustering $n$ data points to $k$ clusters. The convergence proof for the multiplicative updates (\ref{eq:rulesgraph1})-(\ref{eq:rulesgraph2})can be found in the Appendix B.

\section{Experiments}
\label{section3}

In this section we carry out extensive experiments on synthetic and real world data sets to illustrate the effectiveness of the three proposed algorithms: KNSC-Ncut, KNSC-Rcut and KOGNMF. We compare nine recently proposed non-negative spectral clustering algorithms \cite{16,18, 19} and traditional Ncut and Rcut spectral clustering methods \cite{4,2}. Our experimental setting is similar to \cite{16,19}. For the purpose of reproducibility we provide the
code and data sets (see supplementary files).

 
\subsection{Data sets and the evaluation metric}
 We have used the same data sets as in \cite{16, 17, 19}: five UCI \cite{40} data sets and AT\&T face database \cite{38}. The UCI datasets are Soybean, Zoo, Glass, Dermatology and Vehicle. 
The AT\&T face database consists of gray scale face images of 40 persons. Each person has 10 facial images under different light and illumination conditions and the images from the same person belong to the same cluster. 
The important statistics of these data sets are summarized in the Table 2, including the number of samples, dimensions and the number of clusters. 


\begin{table}[!htbp] \centering
	\renewcommand\thetable{2}
  \caption{\textit{Features of the UCI and AT\&T data sets}} 
 \label{tab:data1}
  \scriptsize 
\begin{tabular}{@{\extracolsep{2pt}} lccc} 
\hline 
\hline \\[-1.8ex] 
    Datasets & Samples & Dimension & Clusters \\ \hline  
    Soybean & 47 & 35 & 4 \\ 
    Zoo & 101 & 16 & 7 \\
    AT\&T & 400 & 10304 & 40 \\ 
    Glass & 214 & 9 & 6 \\ 
    Dermatology & 366 & 33 & 6 \\ 
    Vehicle & 846 & 18 & 4 \\
\hline \\[-1.8ex] 
\end{tabular} 
\end{table}

The clustering accuracy is evaluated by the common clustering accuracy measure \cite{16, 18, 19}, which computes the percentage of data points that are correctly clustered with respect to the external ground truth labels. 
For each data point $\textbf{x}_i$ it's label is denoted with $c_i$ and the ground truth cluster index with $g_i$. 
In order to calculate the optimal assignment of labels to cluster indicies $f(c_i)$, the Hungarian bipartite matching algorithm \cite{40} is used, with the complexity $O(k^3)$ for $k$ clusters.
The clustering accuracy can be expressed as:
\begin{equation}
\label{eq:ACC}
ACC = \frac{\sum_{i=1}^n \delta(g_i, f(c_i))}{n},
\end{equation}
where $n$ denotes the total number of data points and the $\delta$ function is defined as
\[
\delta(g_i, c_i)=
\begin{cases} 
      1: & g_i = f(c_i),\\
      0: & g_i \neq f(c_i) .\\
   \end{cases}
   \]

\subsection{Compared algorithms}
We compare our methods to nine recently proposed non-negative spectral clustering approaches and traditional spectral clustering Ncut and Rcut methods:

\begin{itemize}
\item Normalized cut (Ncut) and ratio cut (Rcut) spectral clustering. Ncut spectral clustering exists in different normalizations \cite{4,5}. Our implementation is according to Ncut from  \cite{4}, where eigenvectors of normalized Laplacian matrix $\textbf{Z}$ are normalized such that the $L_2$ norm of each row equals 1.
\item Non-negative spectral clustering methods NSC-Ncut, NSC-Rcut,  and non-negative sparse spectral clustering methods NSSC-Ncut and  NSSC-Rcut from \cite{16}.
\item Global discriminative-based nonnegative spectral clustering methods \cite{19} GDBNSC-Ncut and GDBNSC-Rcut.
\item Symmetric NMF for spectral clustering \cite{18} (NLE). This is the symmetric NMF of the pairwise affinity matrix, which is originally implemented as the standard inner product linear kernel matrix $\textbf{A}  = \textbf{X}^\mathsf{T}\textbf{X}$.

\end{itemize}


\subsection{Clustering results}

We perform $n = 256$ independent runs with random initializations for each of the proposed methods KNSC-Ncut, KNSC-Rcut and KOGNMF. In each run, we randomly initialize matrices ($\textbf{H}, \textbf{Z}, \textbf{F}$) and then iterate multiplicative update rules to achieve convergence and obtain cluster indicator matrix. In all experiments we have used 300 iterations and the convergence occurred after approximately 100 iterations. The cluster memberships for each data point $i$ is obtained by taking the index of the maximal value of $i$-th column in the orthogonal clustering matrix $\textbf{H}$ (or $\textbf{Z}$). For the Rcut and Ncut, the first $k$ eigenvectors are computed once and then 256 runs of \textit{k}-means are performed. 

In Fig. 2 we plot the clustering performance of the NSC-Ncut and KNSC-Ncut on two-dimensional synthetic examples. The synthetic example demonstrates the ability of KNSC-Ncut to separate the nonlinear clusters with high clustering accuracy. In Fig. 3, 4 and 5 we plot the average clustering accuracy over 256 runs on the six data sets. The average clustering accuracy is reported for independent number of runs $2^i$, where $i = 1,2,...,8$. The average clustering accuracy for the Ncut group of algorithms is plotted in the Fig. 3. 
In the Fig. 4 the average clustering accuracy is plotted for the Rcut group.\footnote{The results in Table 3 for the GDBNSC method are reported from original work \cite{19}, however in Fig. 3, 4 and 5 the results for this method were omitted due to the numerical instabilities in reproduction of this method with reported parameters  \cite{19}.} 
 The average clustering accuracy of KOGNMF is shown in Fig.  5. We summarize the average clustering accuracy results for the Ncut and the Rcut group of algorithms in Table 3.
On data sets Dermatology, Glass, Zoo and AT\&T, the KNSC-Ncut clustering accuracy is improved and KNSC-Ncut outperforms Ncut, NSC-Ncut, NSSC-Ncut and GDBNSC-Ncut. On the high dimensional AT\&T face database the clustering accuracy of the KNSC-Ncut algorithm shows considerable improvement.  On  the Soybean and Vehicle data sets the KNSC-Ncut is comparable with the GDBNSC-Ncut. Similary, on Dermatology, Glass, Zoo, Vehicle and AT\&T data set, KNSC-Rcut outperforms Rcut, NSC-Rcut, NSSS-Rcut and GDBNSC-Rcut. In Fig. 5 we plot the average clustering accuracy for the KOGNMF algorithm. The KOGNMF considerably outperforms all algorithms on every data set (Table 3).

\begin{figure*} 
\begin{center}
\includegraphics[width = 0.99\textwidth]{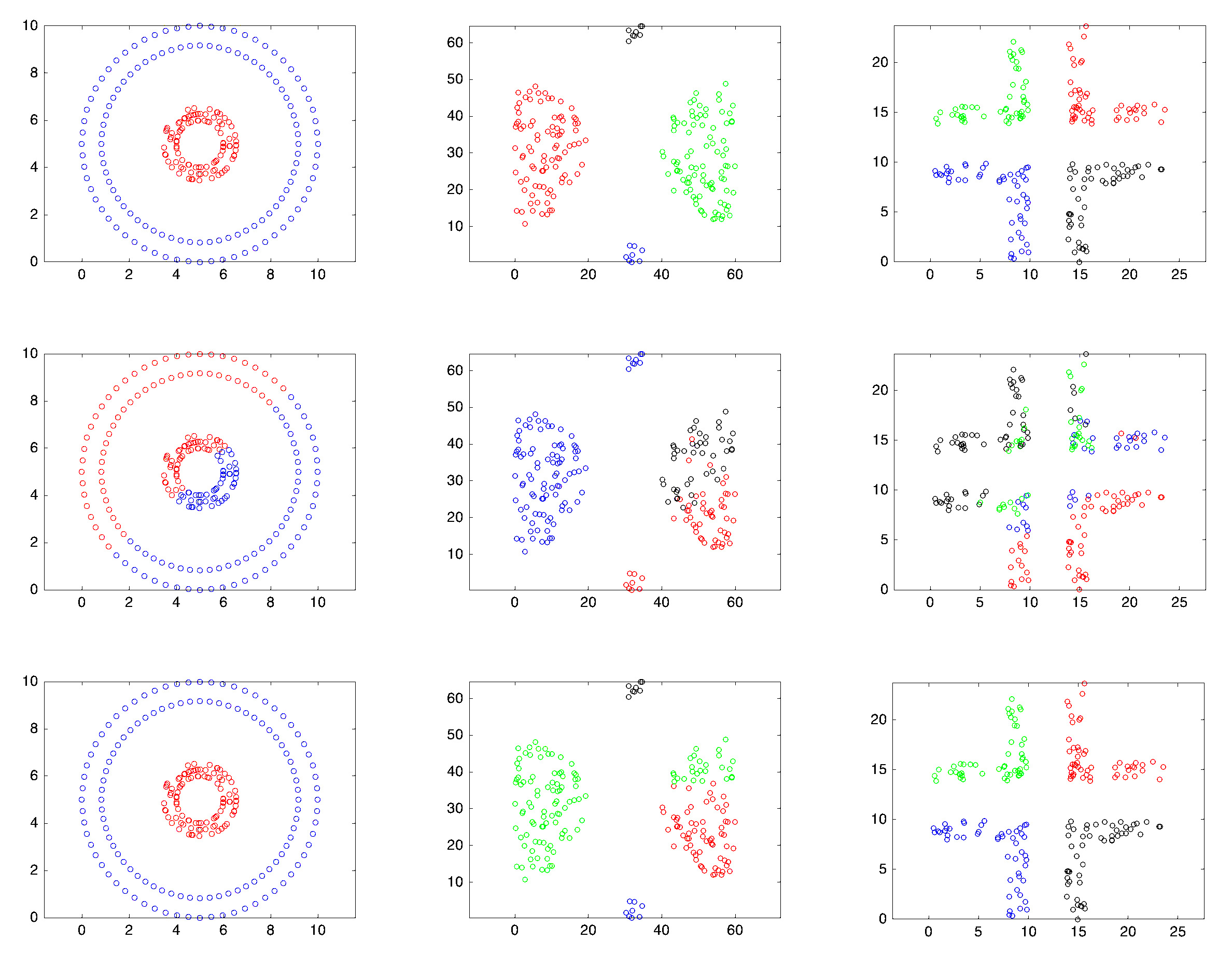} 
\caption{The optimized clustering results of the KNSC-Ncut algorithm comapred with the optimized clustering results of the NSC-Ncut \cite{16}. The two-dimensional data sets with 2 and 4 clusters are plotted in the first row, different clusters are represented with different colors. 
In the second row we plot the clustering results of the NSC-Ncut. The clustering results of the KNSC-Ncut algorithm are plotted in the third row. The clustering accuracy over 256 independent runs is 0.5, 0.7 and 0.62 for NSC-Ncut, and 0.90, 0.85 and 0.82 for the KNSC-Ncut, for the three data sets respectively. The KNSC-Ncut is able to spearate the nonlinear data set composed of two rings of points with high clustering accuracy.
}
\end{center}
\label{fig:sint}
\end{figure*}

\begin{table}[!htbp] \centering
	\renewcommand\thetable{3}
  \caption{\textit{The average clustering accuracy on 5 UCI and AT\&T data sets}} 
 \label{tab:data1}
  \scriptsize 
\begin{tabular}{@{\extracolsep{2pt}} ccccccc} 
\hline 
\hline \\[-1.8ex] 
     & Dermatology & Glass & Soybean & Zoo & Vehicle & AT\&T \\ \hline  
Ncut & 0.75 & 0.46 & 0.70 & 0.63 & 0.37 & 0.62\\
   NSC-Ncut  & 0.71 & 0.25 & 0.71 & 0.61 & 0.39 & 0.35\\ 
     NSSC-Ncut  & 0.71 & 0.34 & 0.71 & 0.66 & 0.41 & 0.02\\
GDBNSC-Ncut  & 0.82 &	0.41 & 0.79 & 0.65 & 0.46 & 0.38\\
\textbf{KNSC-Ncut}    & \textbf{0.87} & \textbf{0.50} & 0.78 & \textbf{0.80} & 0.45 & \textbf{0.70} \\  
 \hline
 Rcut  & 0.47 & 0.41 & 0.63 & 0.60 & 0.33 & 0.31\\
 NSC-Rcut    & 0.66 & 	0.25 &	0.69 & 0.61 & 0.38 & 0.35\\
     NLE     & 0.34  & 0.25 & 0.47 & 0.49  & 0.28 & 0.20 \\
 NSSC-Rcut   & 0.67 & 0.26 & 0.69 & 0.61 & 0.38 & 0.35\\
GDBNSC-Rcut      & 0.73 & 	0.36 & 0.80 & 0.64 & 0.388 & 0.36\\
  \textbf{KNSC-Rcut }      & \textbf{0.87} & \textbf{0.45} & 0.75 & \textbf{0.65} & \textbf{0.45} & \textbf{0.69}  \\
     \hline
\textbf{KOGNMF}  & \textbf{0.91} & \textbf{0.48} & \textbf{0.80} & \textbf{0.78} & \textbf{0.45} & \textbf{0.70}\\  \hline
\hline \\[-1.8ex] 
\end{tabular} 
\caption{The average clustering accuracy of KNSC-Ncut, KNSC-Rcut and KOGNMF compared with 9 recently proposed NMF-based NSC methods on the 5 UCI \cite{40} data sets and the AT\&T face database \cite{38}. KNSC-Rcut performs considerably better on 4 data sets, and has a comparable accuracy on two data sets. KNSC-Nuct algorithm outperforms on 5 data sets, and has a comparable clustering accuracy on one data set. KOGNMF algorithm has considerably better accuracy on 4 data sets, including the difficult AT\&T face images database, and is comparable on two data sets. All three algorithms have considerably higher clustering accuracy on the difficult AT\&T face database.}
\end{table}


\begin{figure}
        \begin{subfigure}[b]{0.5\textwidth}
                \includegraphics[width=\linewidth]{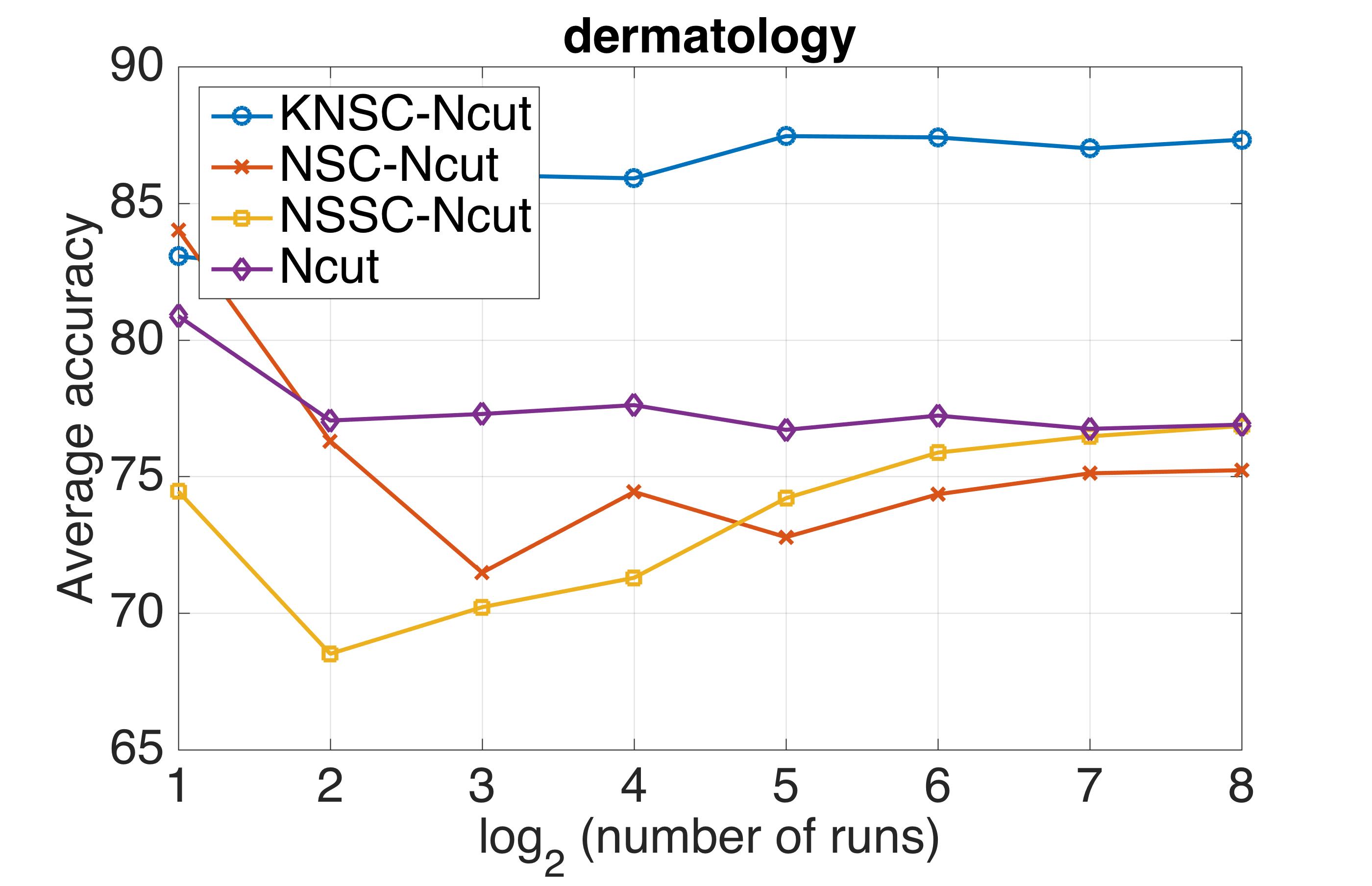}
        \end{subfigure}%
        \begin{subfigure}[b]{0.5\textwidth}
                \includegraphics[width=\linewidth]{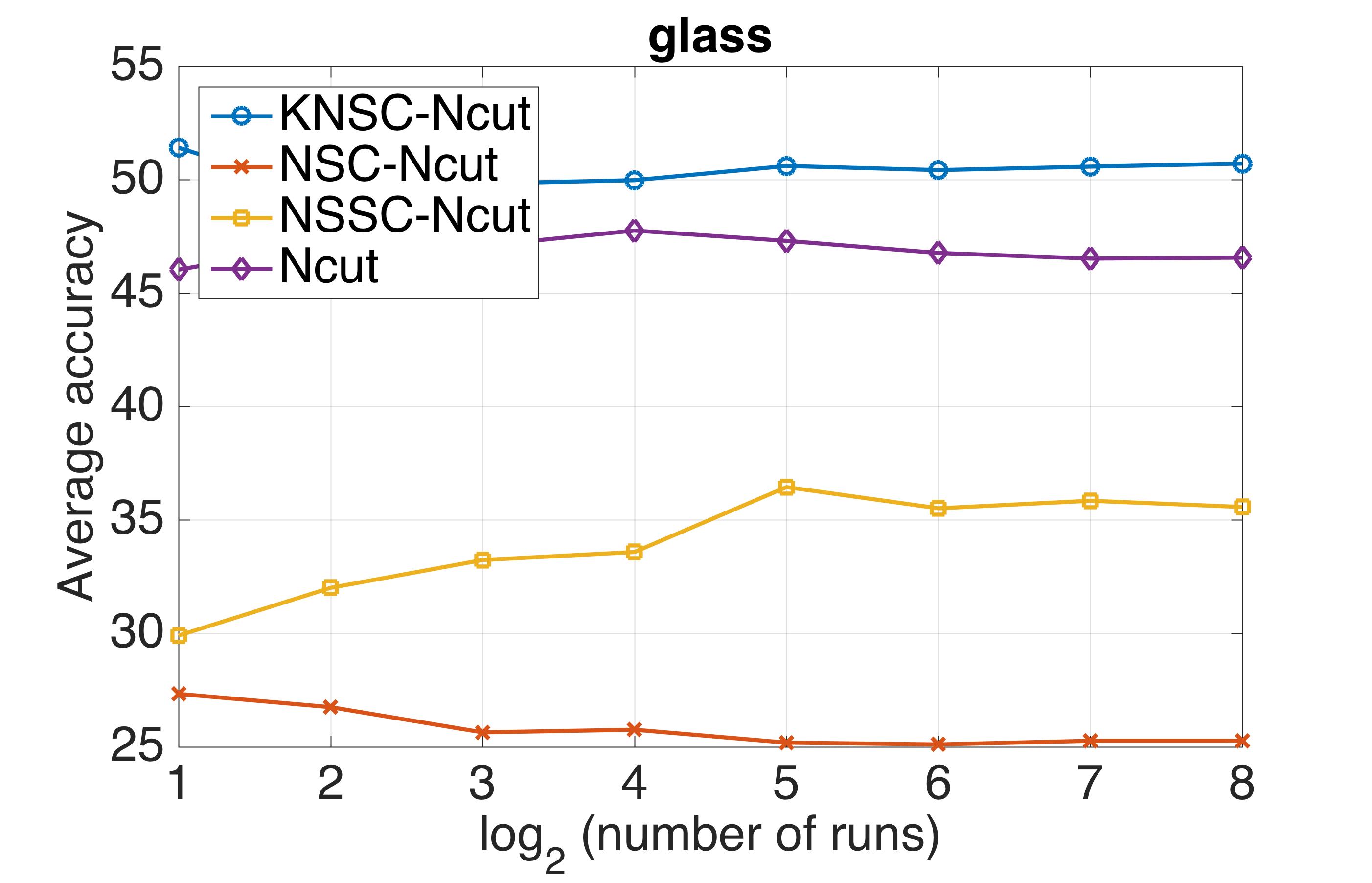}
        \end{subfigure}
        \begin{subfigure}[b]{0.5\textwidth}
                \includegraphics[width=\linewidth]{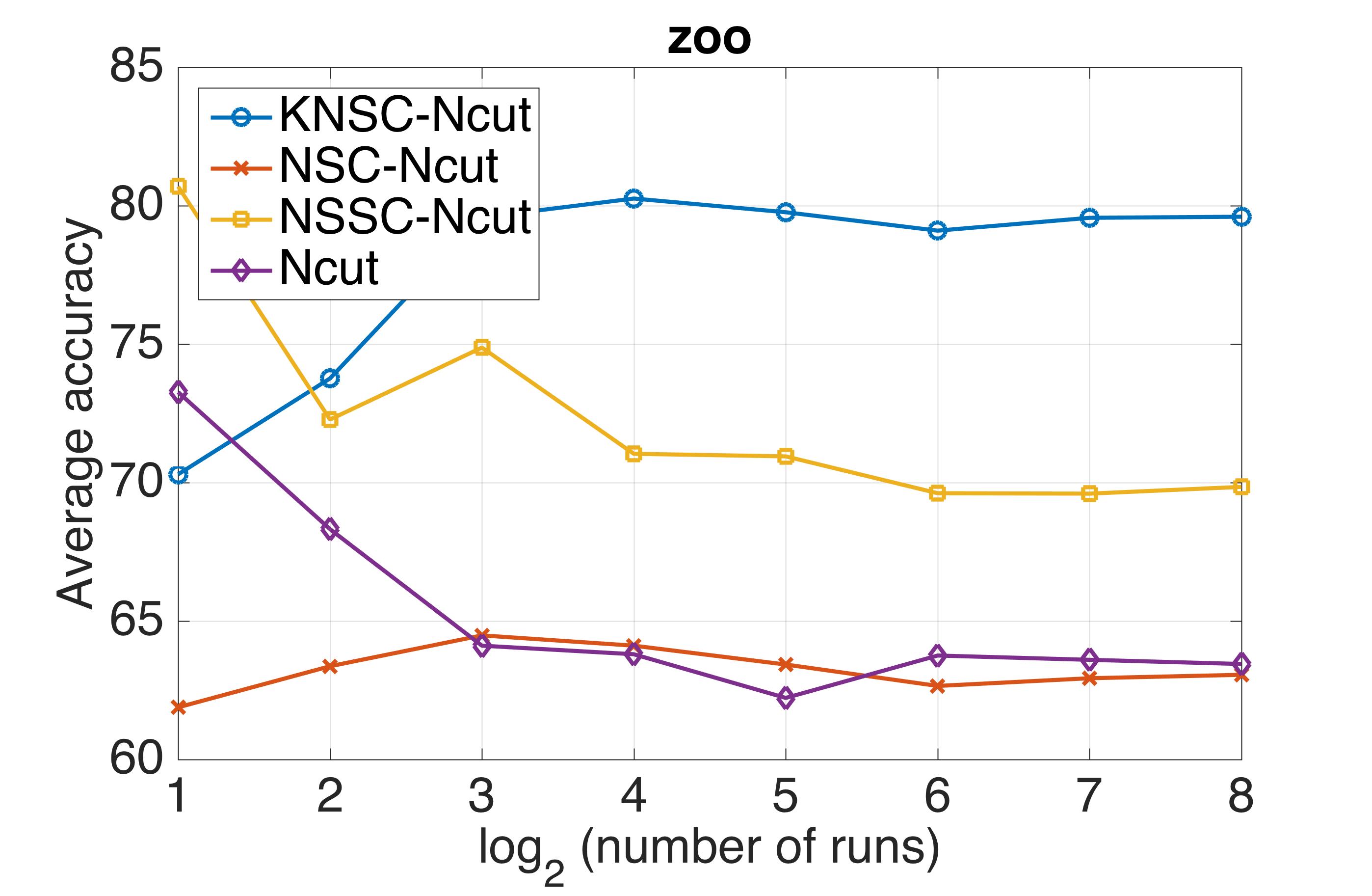}
        \end{subfigure}%
        \begin{subfigure}[b]{0.5\textwidth}
                \includegraphics[width=\linewidth]{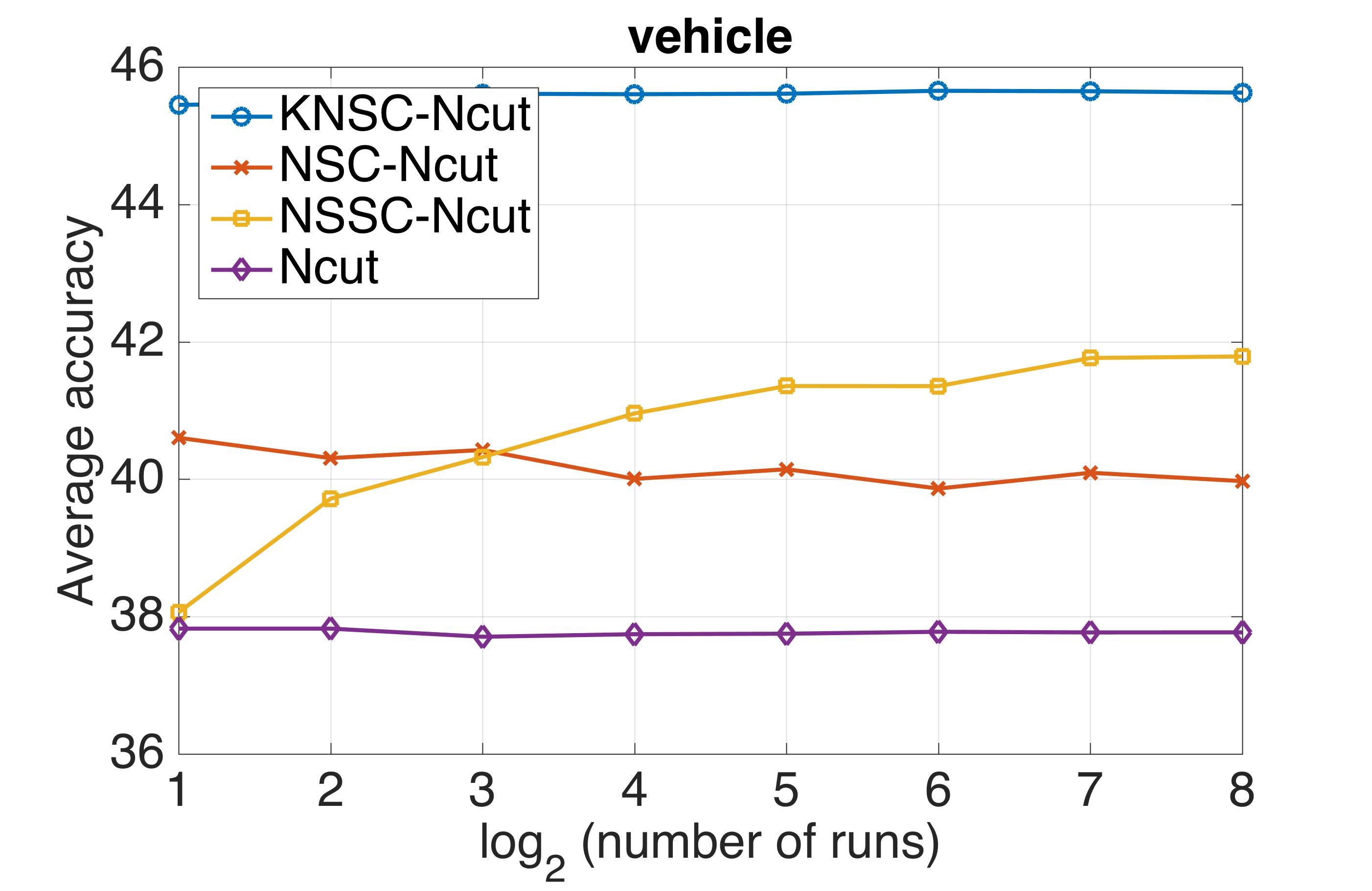}
        \end{subfigure}
        \begin{subfigure}[b]{0.5\textwidth}
                \includegraphics[width=\linewidth]{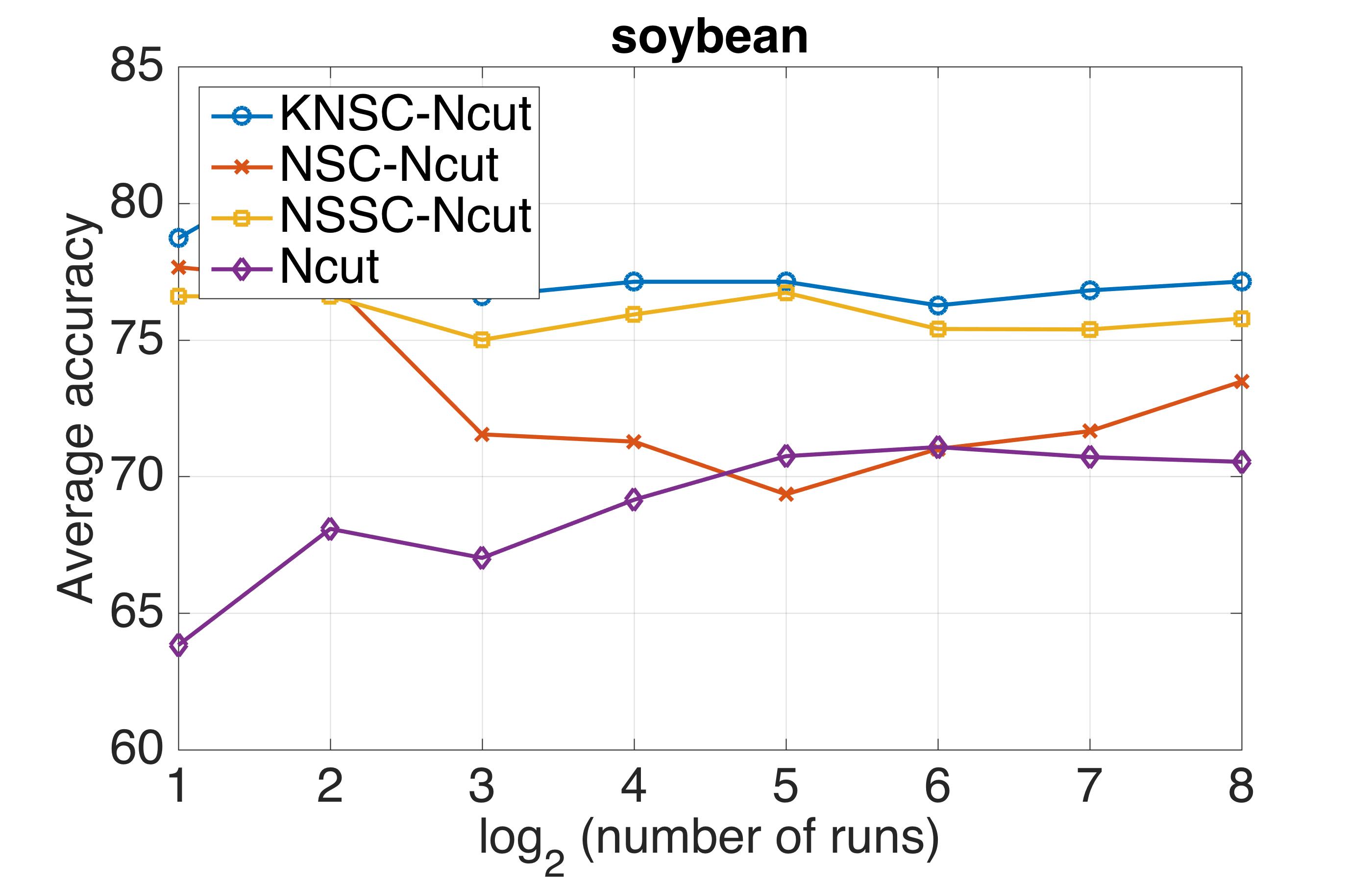}
        \end{subfigure}%
        \begin{subfigure}[b]{0.5\textwidth}
                \includegraphics[width=\linewidth]{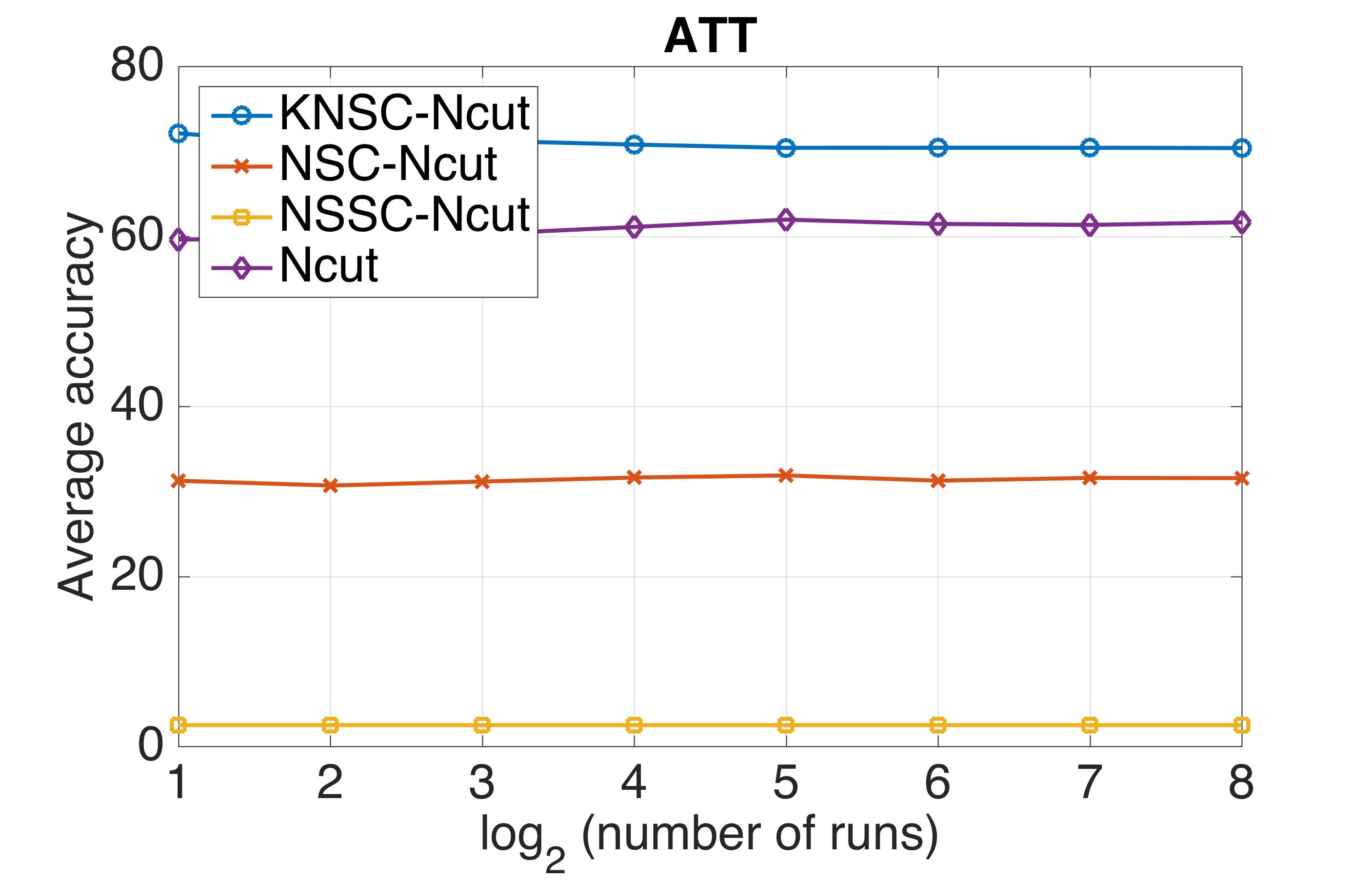}
        \end{subfigure}
        \caption{The average clustering accuracy of KNSC-Ncut algorithm compared with Ncut, NSC-Ncut and NSSC-Ncut algorithms on five UCI \cite{40} data sets and AT\&T face database \cite{38}. 
The average clustering accuracy is plotted for the independent number of runs $2^i = \{ 2,4, ..., 256 \}$. The clustering accuracy of KNSC-Ncut is higher on the majority of data sets. The clustering accuracy for the AT\&T face database is considerably improved when compared with the state-of-the-art non-negative spectral clustering methods.}
\end{figure}


\begin{figure}
        \begin{subfigure}[b]{0.5\textwidth}
                \includegraphics[width=\linewidth]{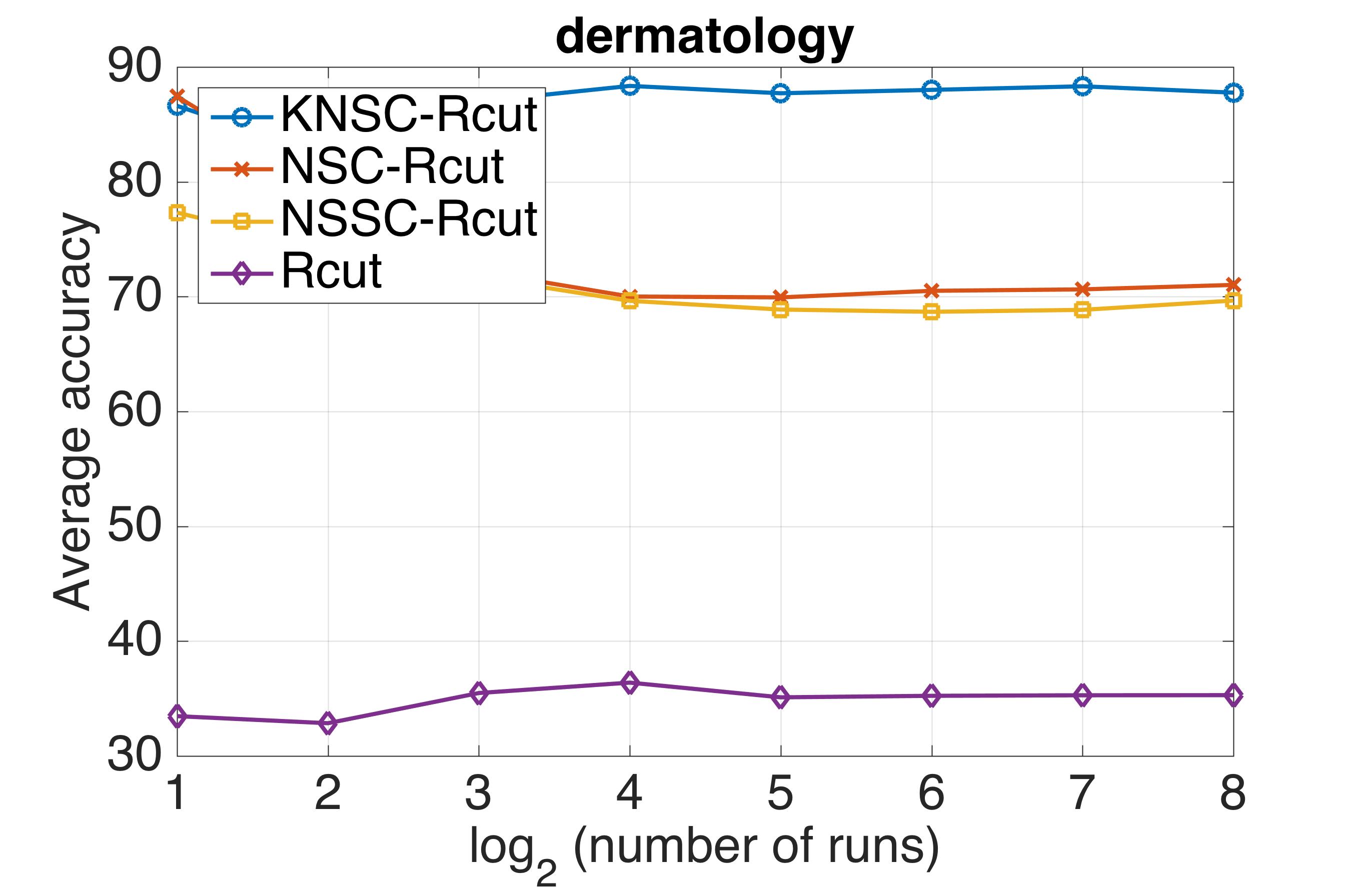}
        \end{subfigure}%
        \begin{subfigure}[b]{0.5\textwidth}
                \includegraphics[width=\linewidth]{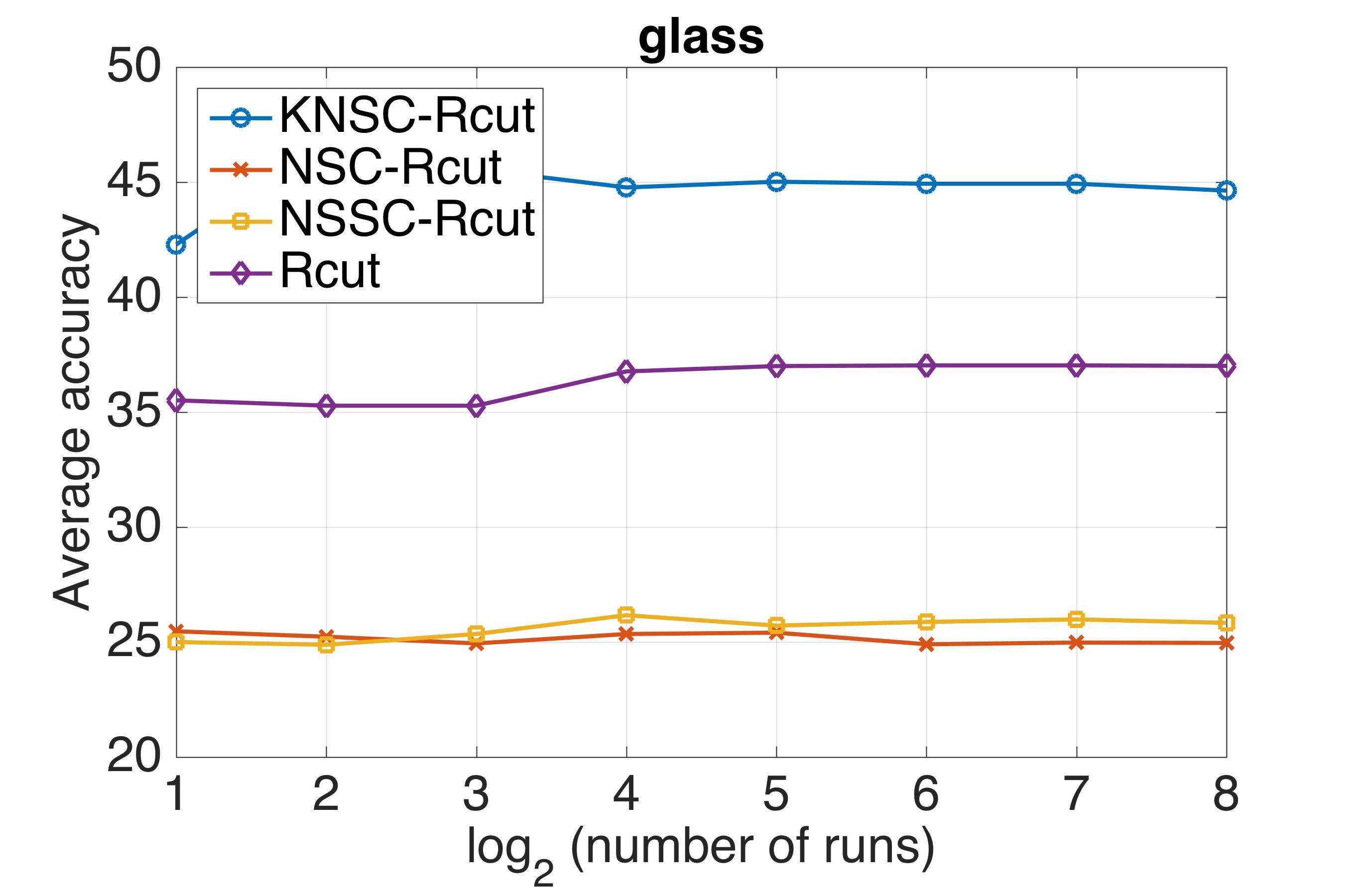}
        \end{subfigure}
        \begin{subfigure}[b]{0.5\textwidth}
                \includegraphics[width=\linewidth]{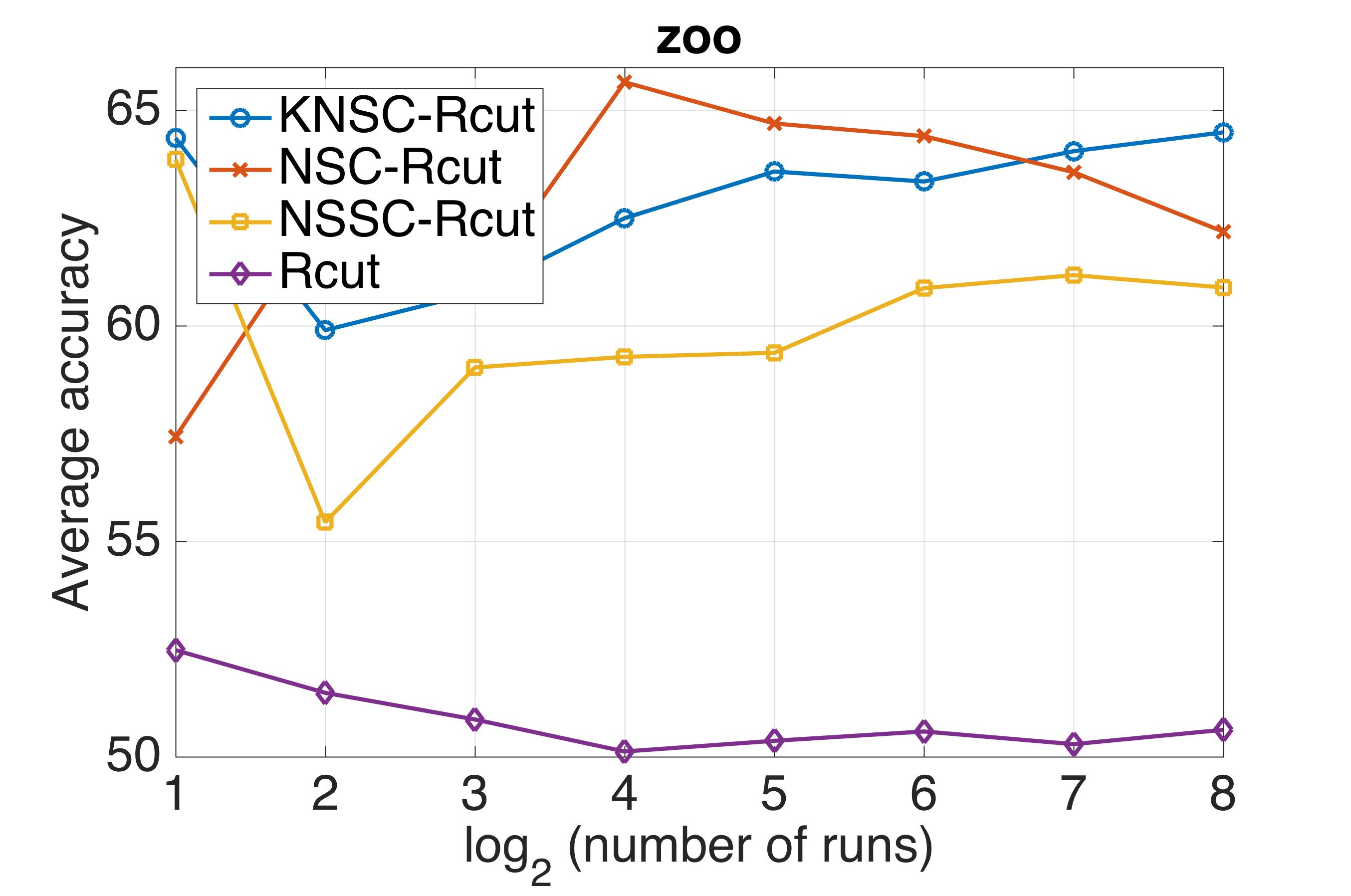}
        \end{subfigure}%
        \begin{subfigure}[b]{0.5\textwidth}
                \includegraphics[width=\linewidth]{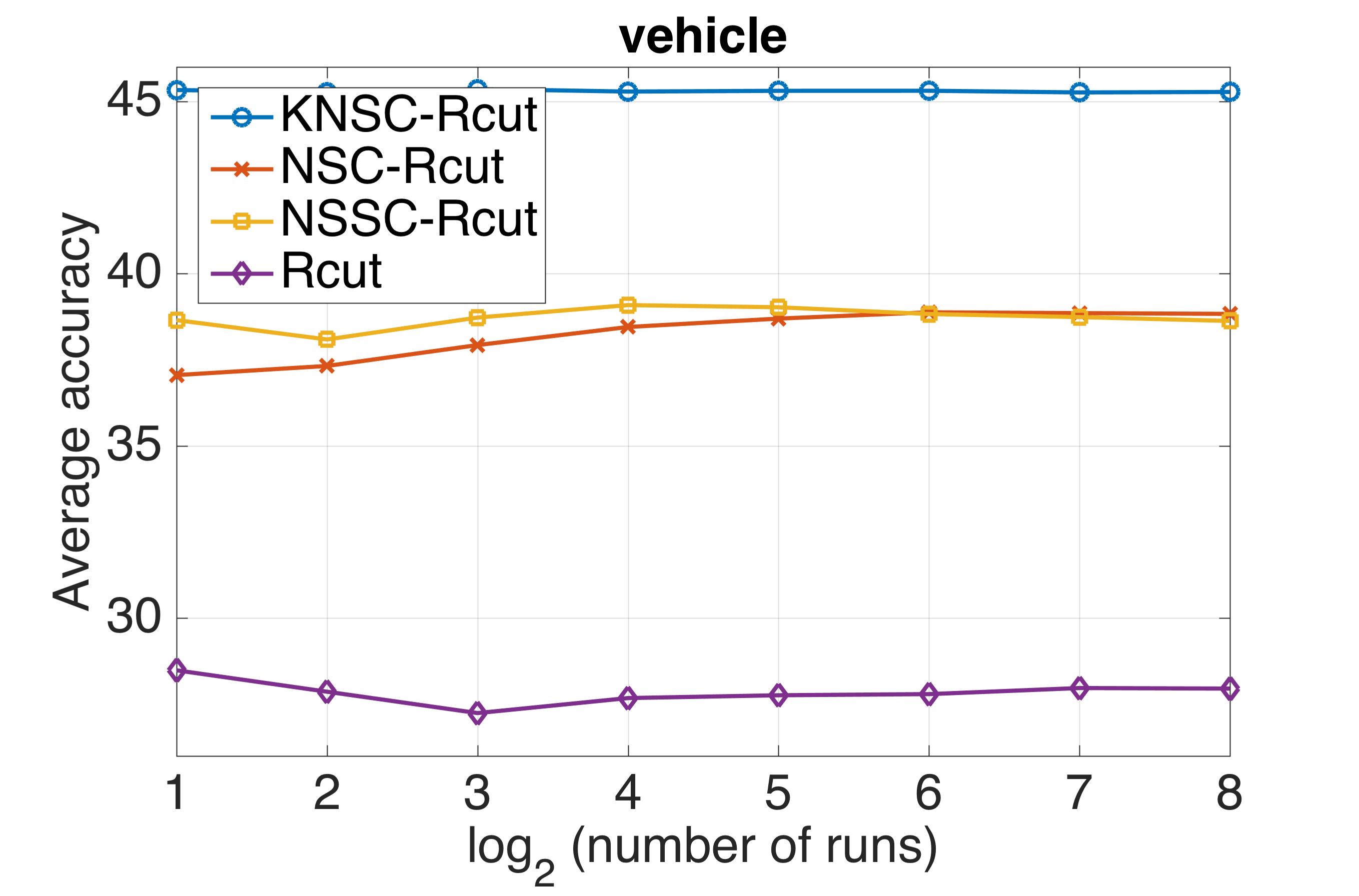}
        \end{subfigure}
        \begin{subfigure}[b]{0.5\textwidth}
                \includegraphics[width=\linewidth]{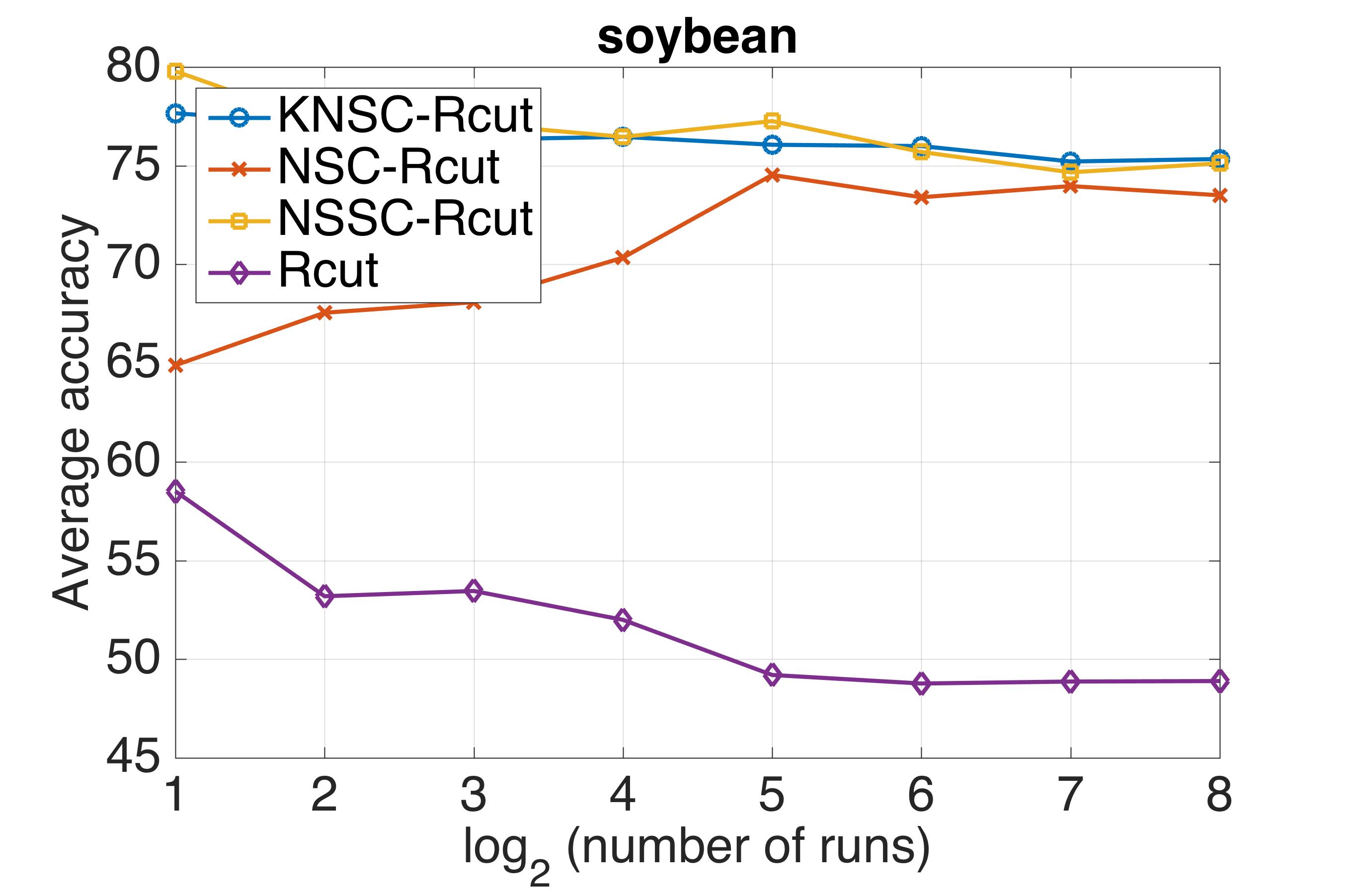}
        \end{subfigure}%
        \begin{subfigure}[b]{0.5\textwidth}
                \includegraphics[width=\linewidth]{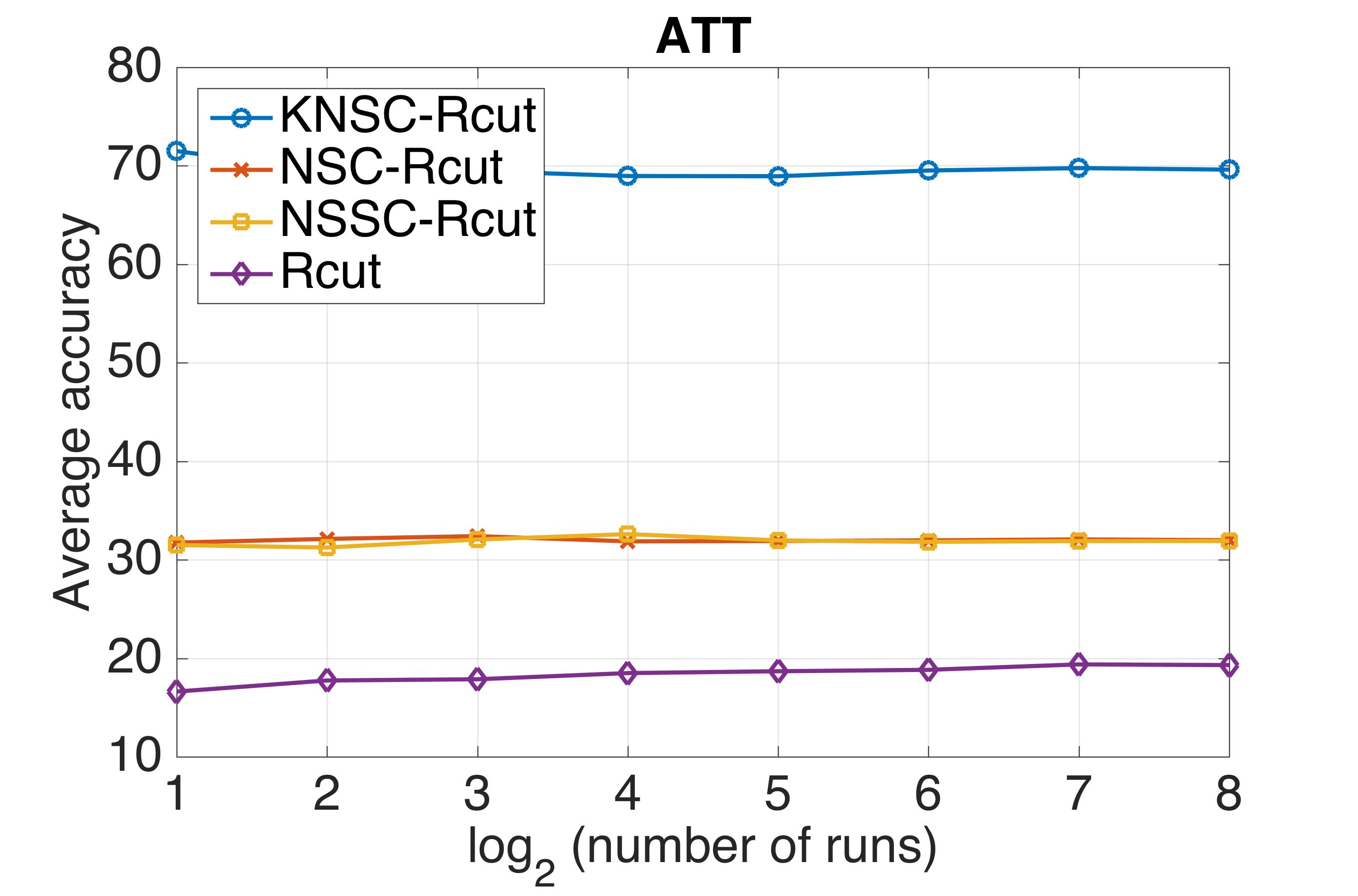}
        \end{subfigure}
        \caption{The average clustering accuracy of KNSC-Rcut algorithm compared with Rcut, NSC-Rcut and NSSC-Rcut algorithms on five UCI \cite{40} data sets and AT\&T face database \cite{38}. 
The average clustering accuracy is plotted for the independent number of runs $2^i = \{ 2,4, ..., 256 \}$.  
The KNSC-Rcut algorithm outperforms NSC algorithms on the majority of data sets.}
\end{figure}


\begin{figure}
        \begin{subfigure}[b]{0.5\textwidth}
                \includegraphics[width=\linewidth]{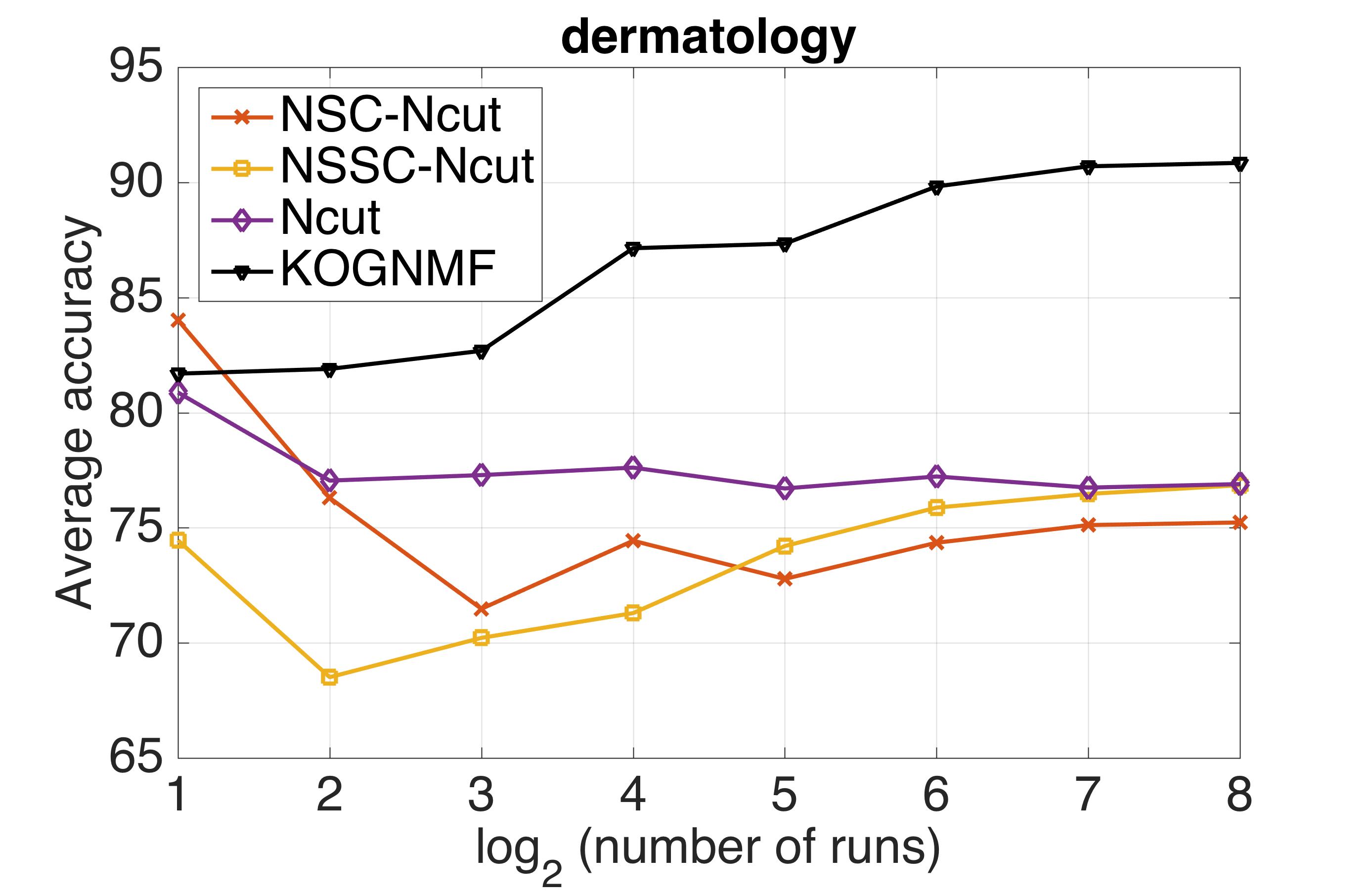}
        \end{subfigure}%
        \begin{subfigure}[b]{0.5\textwidth}
                \includegraphics[width=\linewidth]{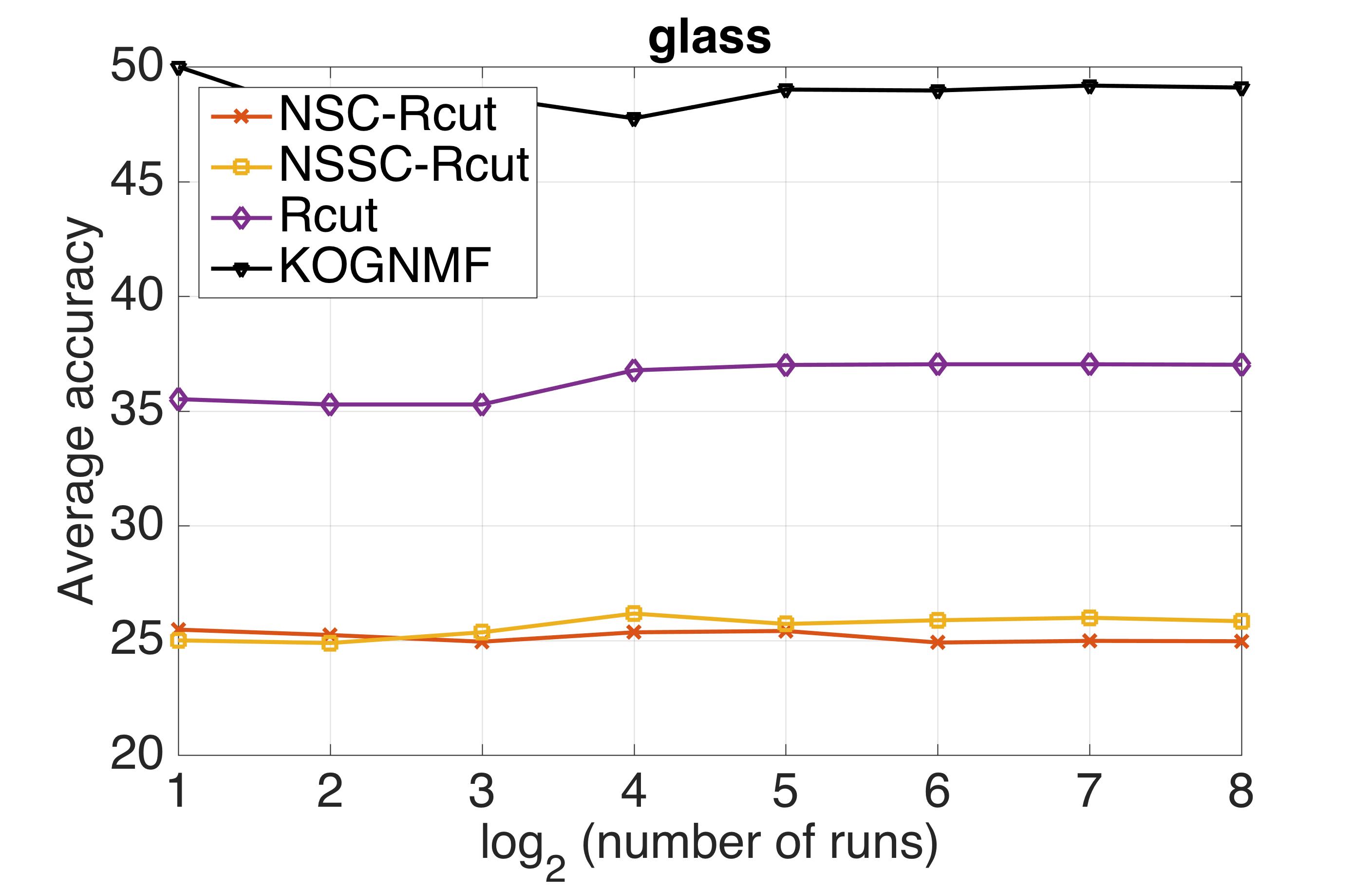}
        \end{subfigure}
        \begin{subfigure}[b]{0.5\textwidth}
                \includegraphics[width=\linewidth]{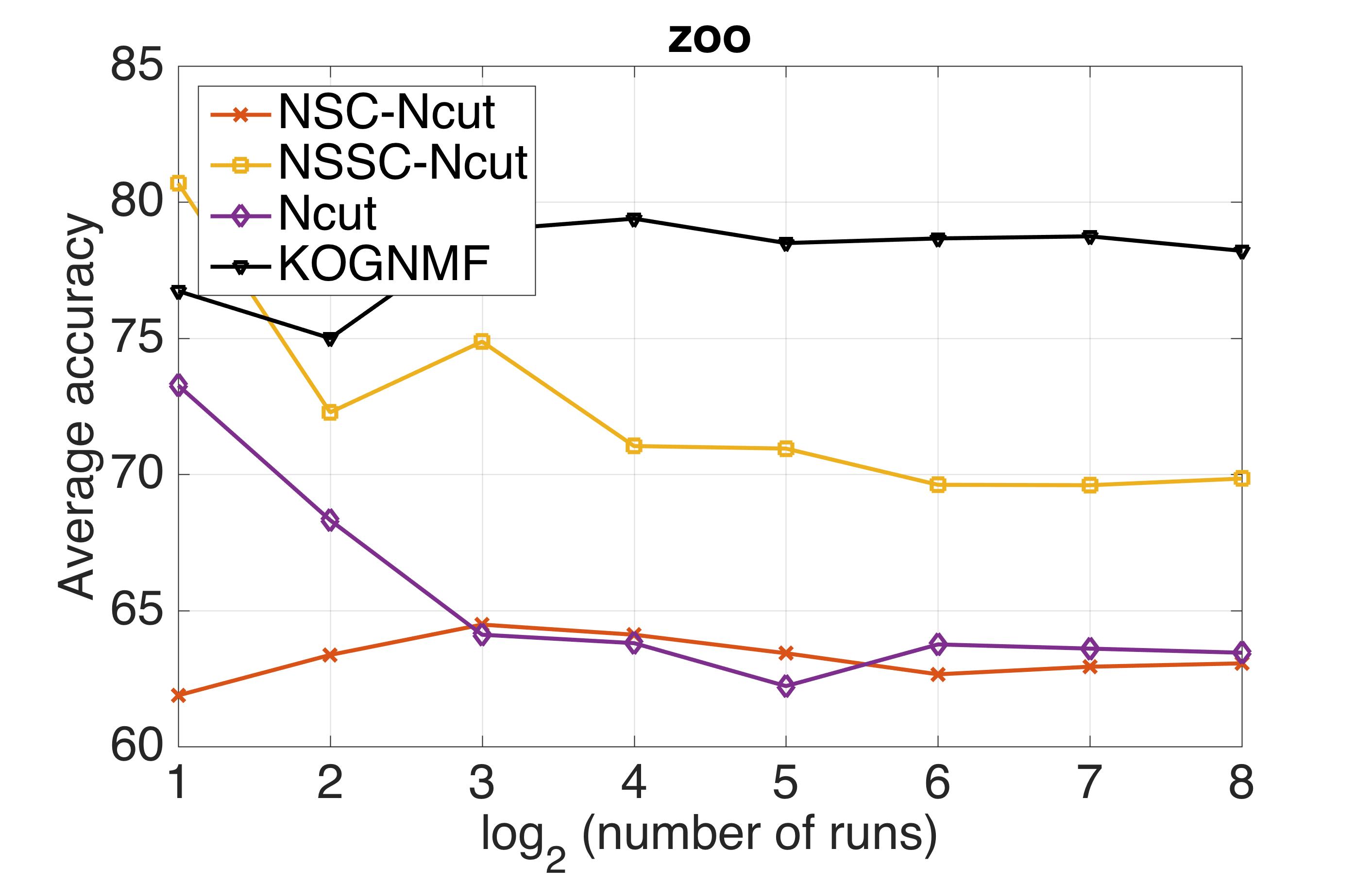}
        \end{subfigure}%
        \begin{subfigure}[b]{0.5\textwidth}
                \includegraphics[width=\linewidth]{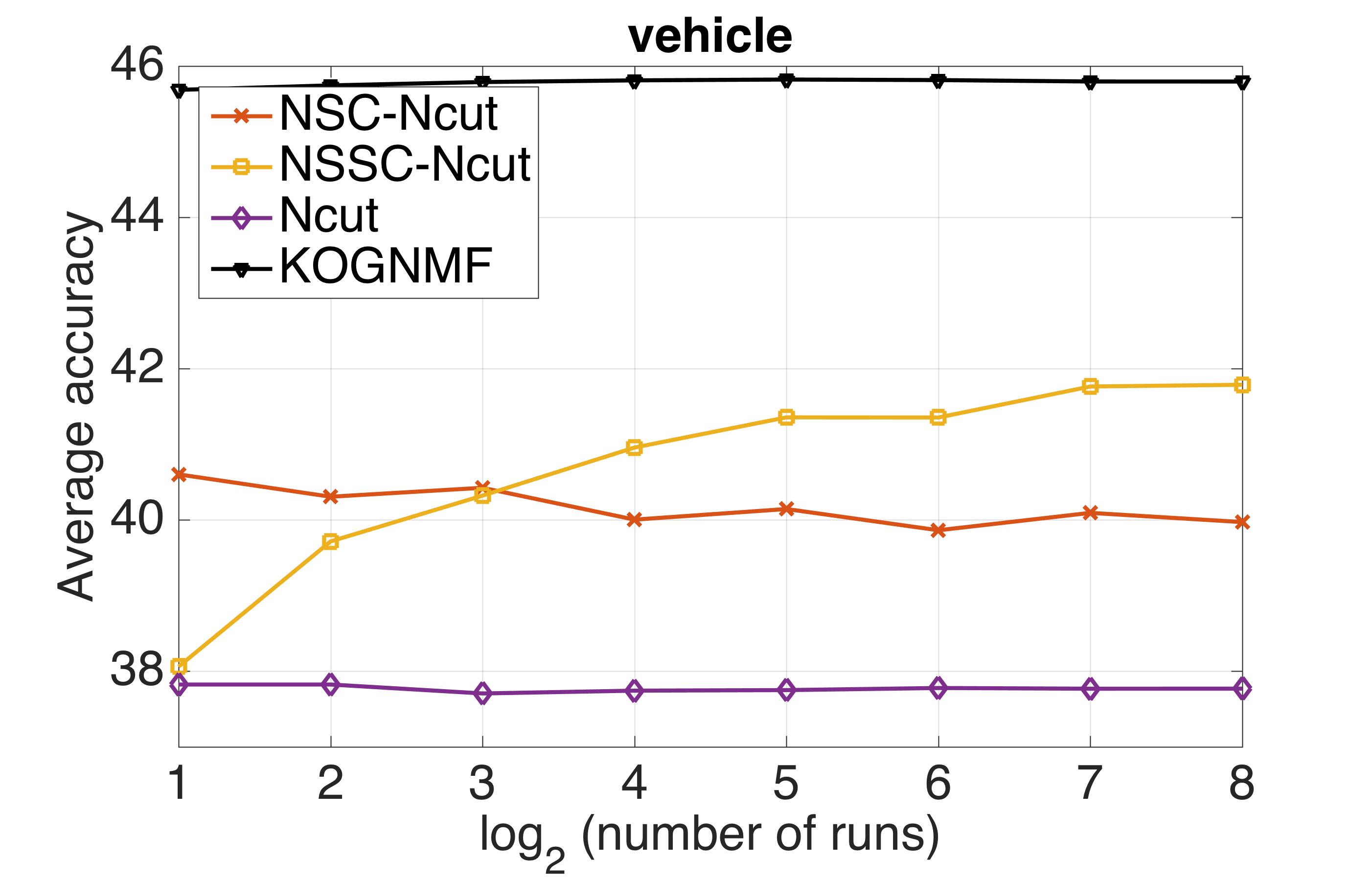}
        \end{subfigure}
        \begin{subfigure}[b]{0.5\textwidth}
                \includegraphics[width=\linewidth]{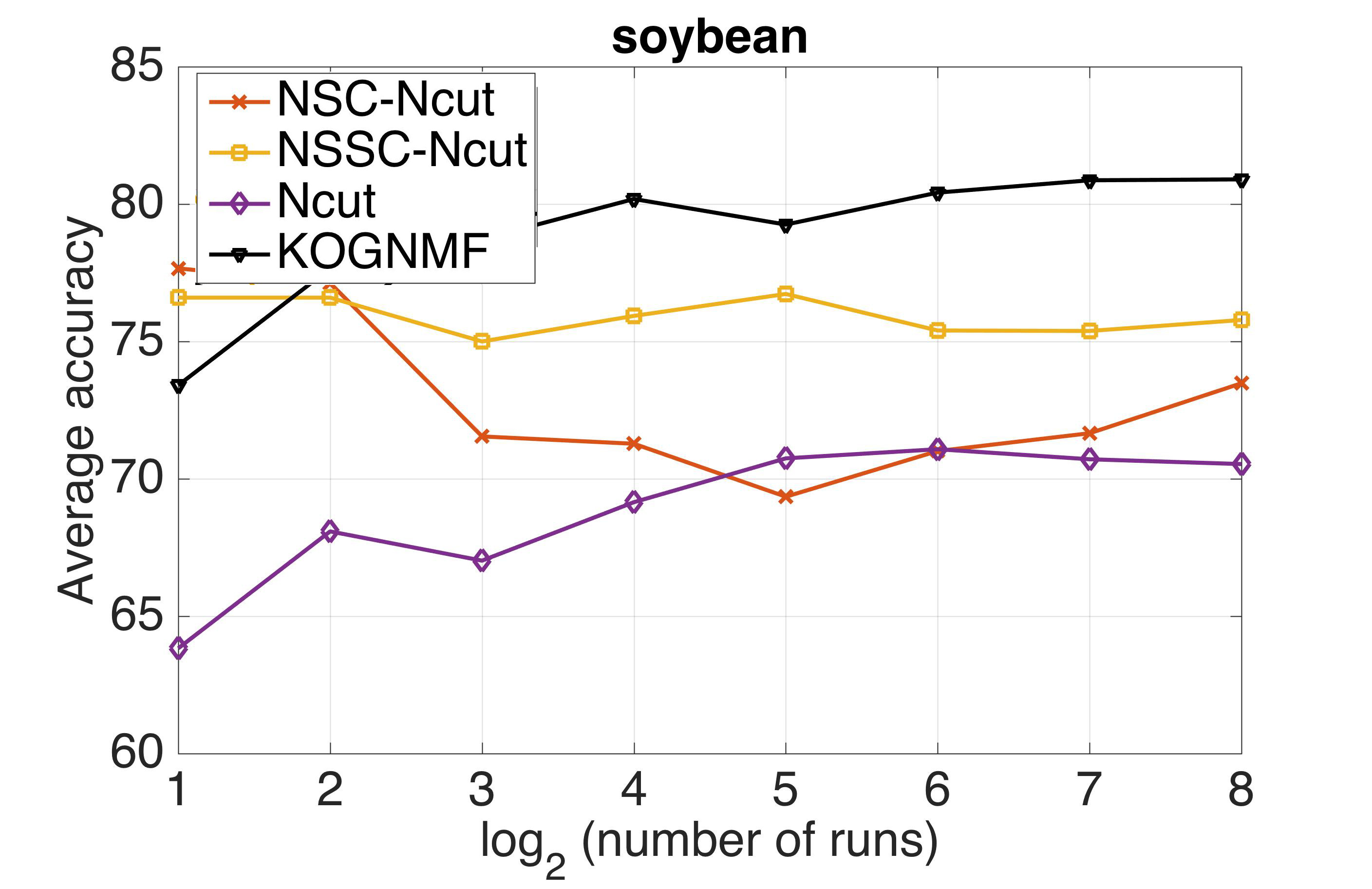}
        \end{subfigure}%
        \begin{subfigure}[b]{0.5\textwidth}
                \includegraphics[width=\linewidth]{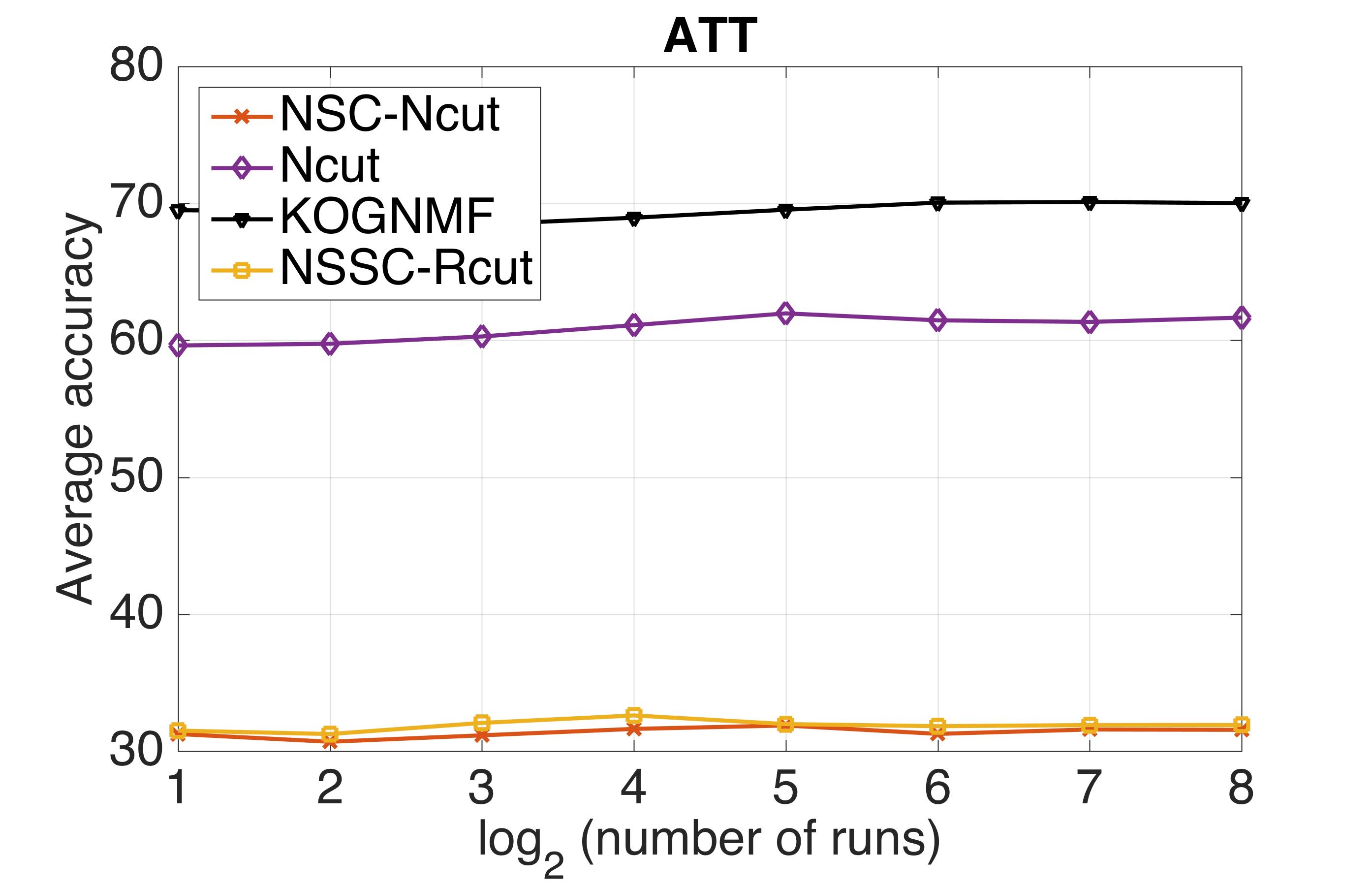}
        \end{subfigure}
        \caption{The average clustering accuracy of KOGNMF algorithm on 5 UCI \cite{40} data sets and AT\&T face database. The average clustering accuracy is plotted for the independent number of runs $2^i = \{ 2,4, ..., 256 \}$. The KOGNMF algorithm outperforms all non-negative spectral clustering methods on every data set, including the difficult AT\&T face database \cite{38}.}
\end{figure}


\begin{figure}
        \begin{subfigure}[b]{0.5\textwidth}
                \includegraphics[width=\linewidth]{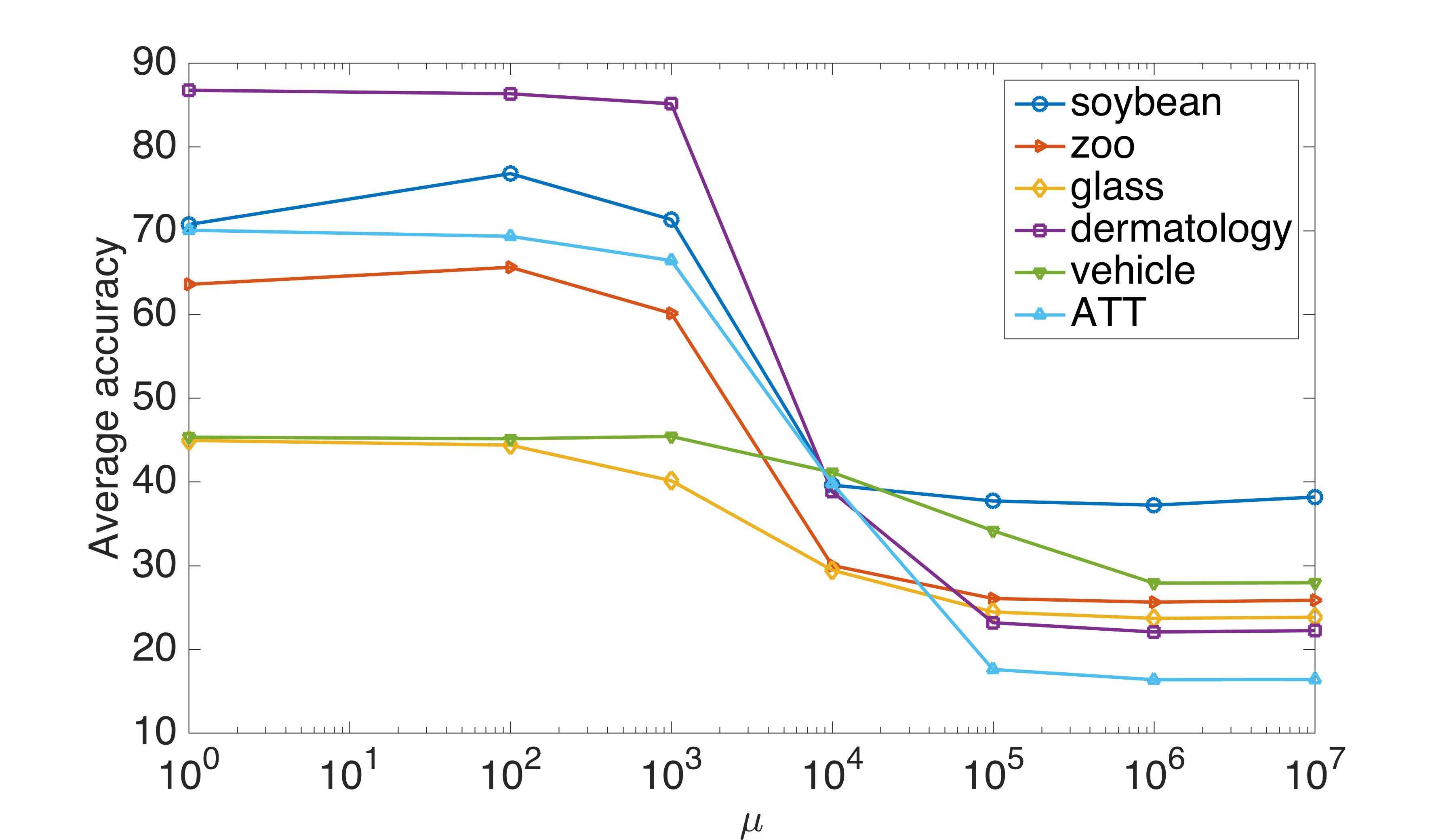}
        \end{subfigure}%
        \begin{subfigure}[b]{0.5\textwidth}                \includegraphics[width=\linewidth]{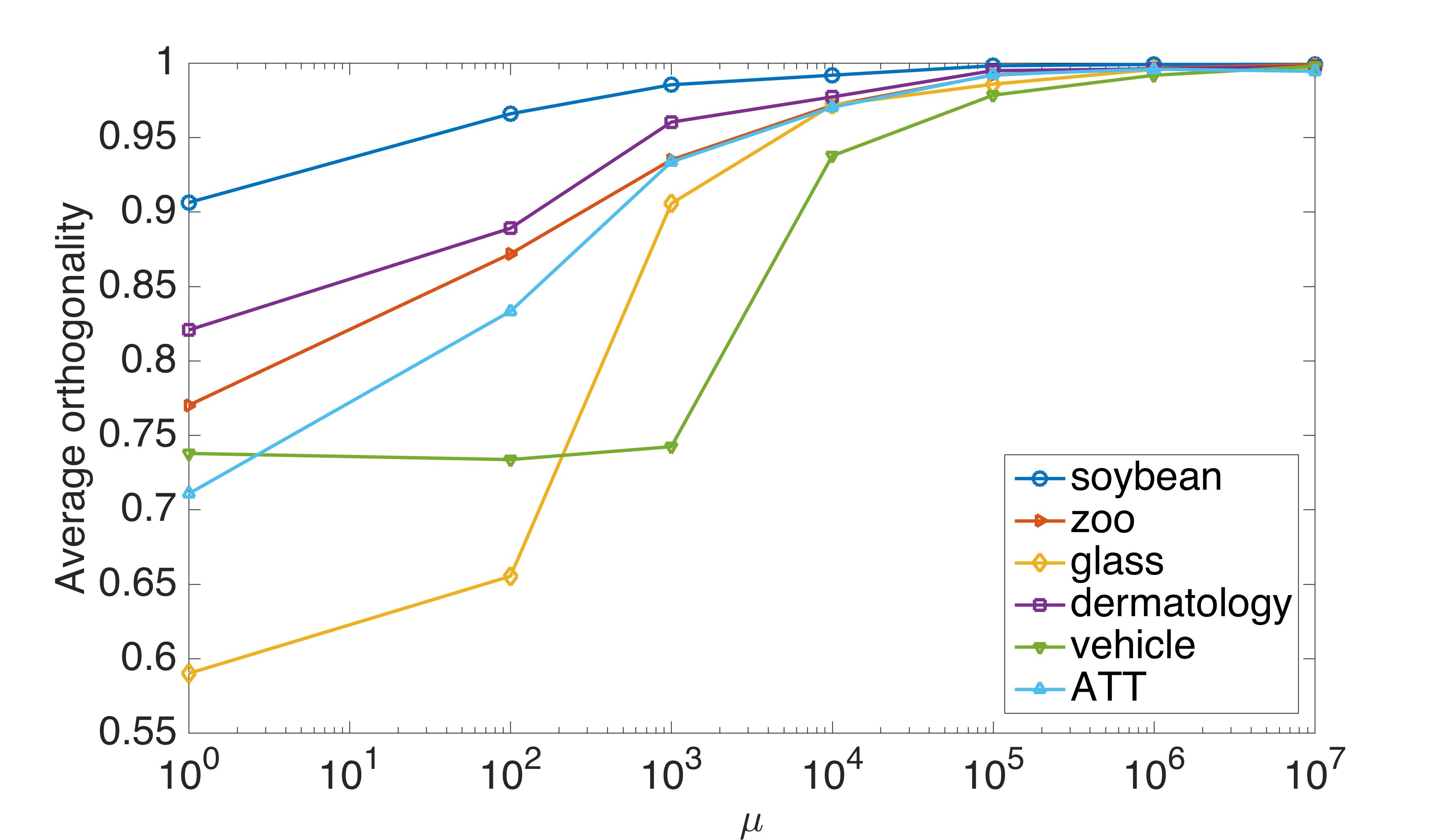}
        \end{subfigure}
        \caption{Left: The average orthogonality of the clustering matrix $\textbf{H}$ (KNSC-Rcut) over the 256 runs, plotted for fixed reconstruction error parameter $\alpha=10$ and for a wide range of values of the orthogonality parameter $\mu$ on all six data sets.  Right: The average clustering accuracy of KNSC-Rcut for fixed $\alpha=10$ plotted for different values of the parameter $\mu$.  The average orthogonality of the clustering matrix $\textbf{H}$ increases up to 1 if the parameter $\mu$ is increased. The average clustering accuracy is robust for all six data sets for a wide range of the trade-off parameter $\mu$.}
\end{figure}


\begin{figure}
        \begin{subfigure}[b]{0.5\textwidth}
                \includegraphics[width=\linewidth]{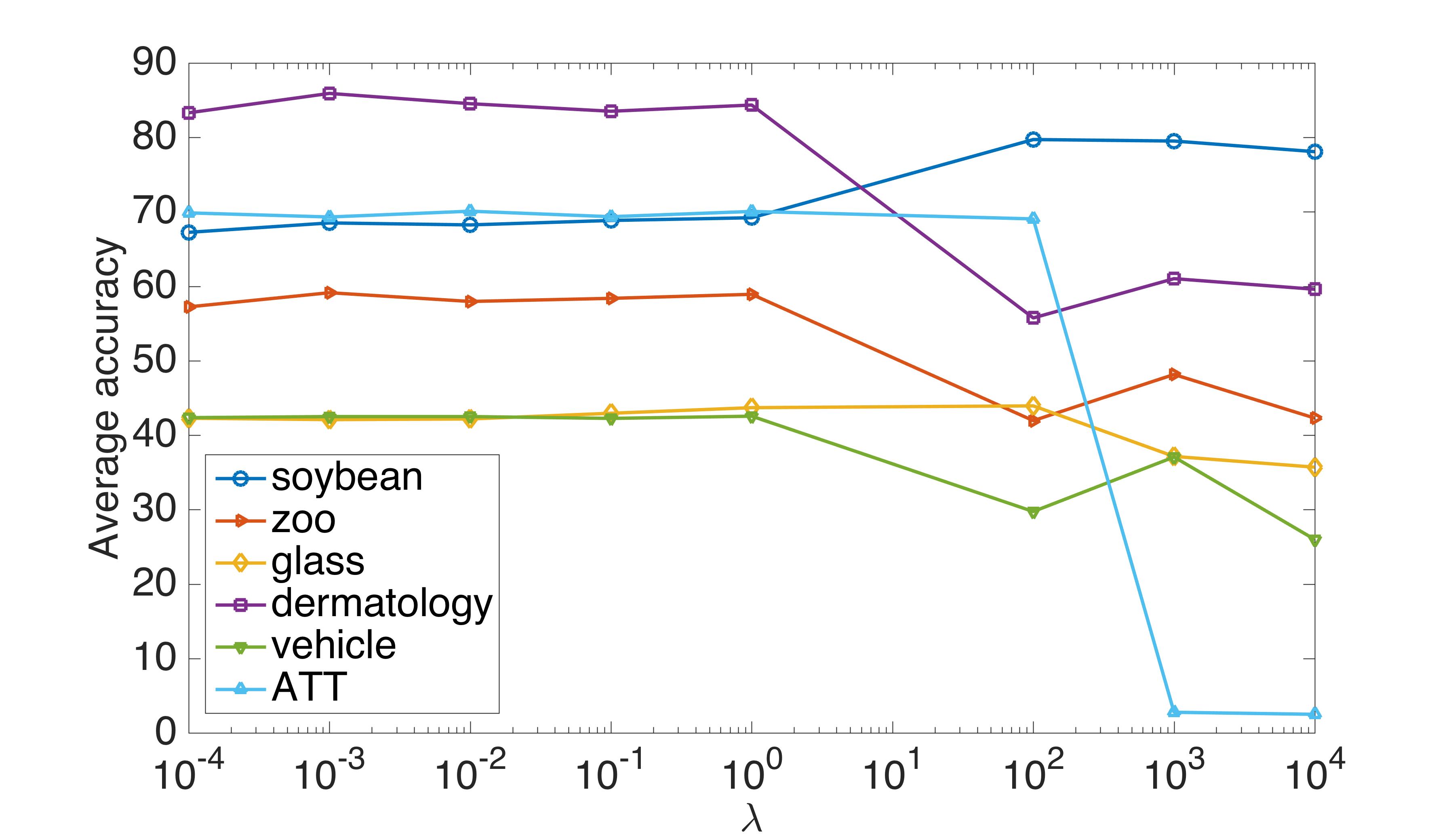}
        \end{subfigure}%
        \begin{subfigure}[b]{0.5\textwidth}                \includegraphics[width=\linewidth]{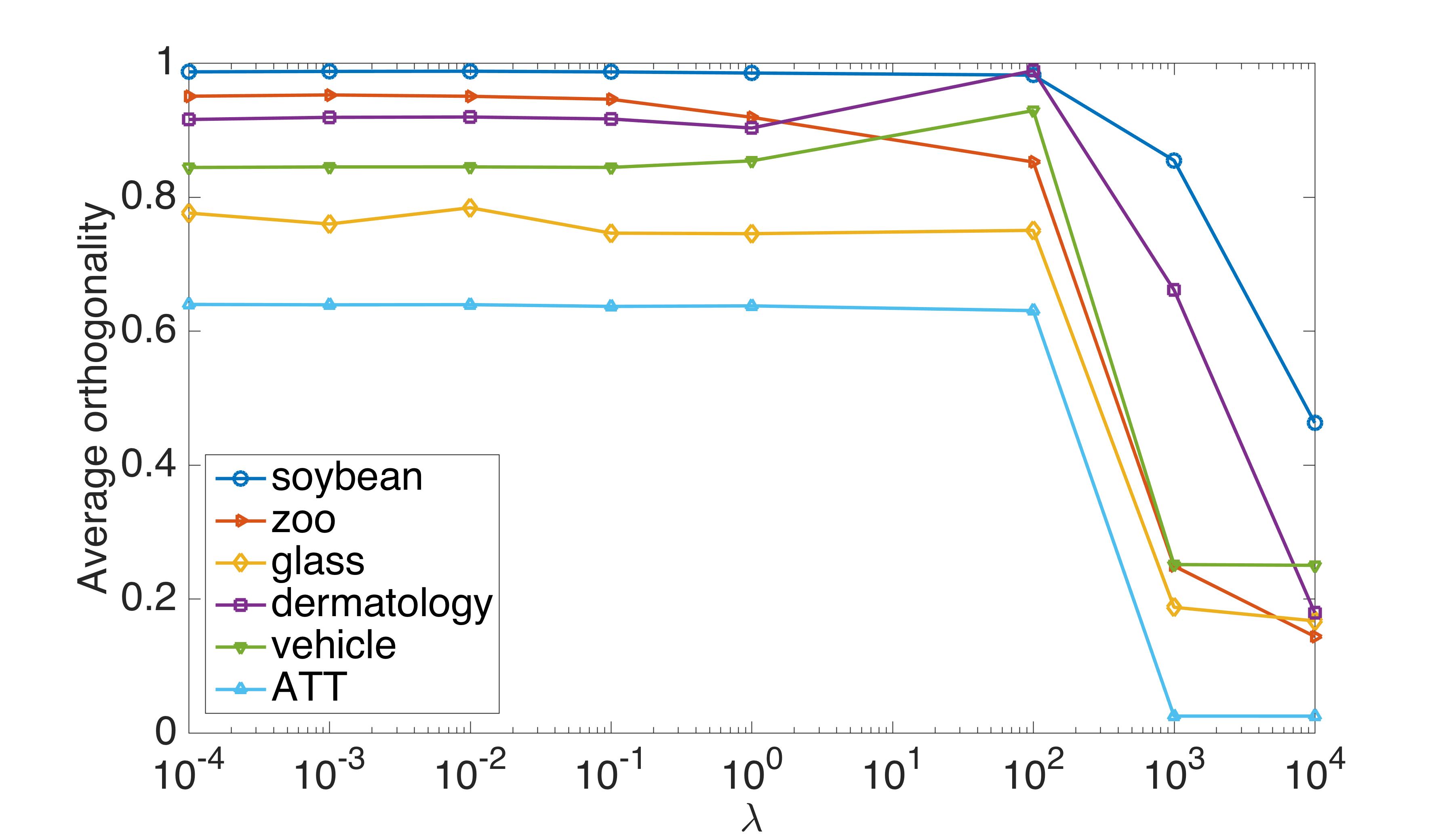}
        \end{subfigure}
        \caption{Left: The average orthogonality of the clustering matrix $\textbf{H}$ (KNSC-Rcut) over the 256 runs, plotted for fixed reconstruction error parameter $\alpha=10$ and orthogonality regularization parameter $\mu=100$ for different values of the graph regularization parameter $\lambda$ on all six data sets. Right: The average clustering accuracy of KNSC-Rcut for fixed parameters $\alpha=10$ and $\mu=100$ plotted for different values of the parameter $\lambda$. The average clustering accuracy is robust for all six data sets for a wide range of the trade-off parameter $\lambda$.}
\end{figure}

\begin{table}[!htbp] \centering
	\renewcommand\thetable{4}
  \caption{\textit{The average clustering accuracy on the hold-out validation set}} 
\label{tab:data3}
  \scriptsize 
\begin{tabular}{@{\extracolsep{2pt}} ccccccc} 
\hline 
\hline \\[-1.8ex] 
  \hline
    Datasets & Dermatology & Glass & Soybean & Zoo & Vehicle & AT\&T \\ \hline  
    \multirow{1}{*}{  NLE  }
    &  0.37 & 0.38 & 0.55 & 0.45 & 0.33 & 0.26\\ 
     \multirow{1}{*}{   KNSC-Ncut }
      & \textbf{0.87} & \textbf{0.47} & 0.73 & \textbf{0.77} & \textbf{0.47} & \textbf{0.70} \\  
   \multirow{1}{*}{  KNSC-Rcut }     
     & \textbf{0.85} & \textbf{0.47} & 0.76 & \textbf{0.67} & \textbf{0.48} & \textbf{0.73} \\   
     \multirow{1}{*}{ KOGNMF }
       & \textbf{0.89} & \textbf{0.49} & \textbf{0.76} & \textbf{0.78} & \textbf{0.48} & \textbf{ 0.73} \\ 
\hline \\[-1.8ex] 
\end{tabular} 
\caption{The hold-out validation consists of randomly splitting each data set into two equally sized parts with the equally distributed cluster membership. The grid search optimization is performed on the first half of the data set, while the second half is used as a hold-out validation where optimized parameters are used. For each data set, we measure the average score over 256 independent runs on the hold-out data. We denote with bold our results that outperform the optimized clustering accuracy scores of the state-of-the-art NSC methods without the hold-out validation. The KNSC-Ncut and KNSC-Rcut algorithms have higher average clustering accuracy on the majority of data sets, while KOGNMF algorithm outperforms on all six data sets.}
\end{table}

\subsection{The parameter selection}

The kernel-based orthogonal NMF multiplicative rules have in total four parameters: $\alpha$, $\mu$ and $\lambda$ and the Gaussian kernel width $\sigma$. The three parameters $\alpha$, $\mu$ and $\lambda$ are a trade-off parameters which balance the reconstruction error, orthogonality regularization and the graph regularization, respectively. In all the experiments and data sets we have fixed the three trade-off parameters to the same constant values $\alpha = 10$, $\mu = 100$ and $\lambda = 10$. Furthermore, the three trade-off parameters can be reduced to two, as the NMF objective functions given in the Eqs. (\ref{nonlinmin}) and (\ref{nonlinmingr}) can be divided by $\alpha$. By fixing the trade-off parameters throughout all of the experiments we effectively need to optimize only one parameter, which is the kernel width. For the trade-off parameters we perform sensitivity analysis to demonstrate that the constant values of the trade-off parameters can be chosen in a wide range of values (a few orders of magnitude), as shown in Fig. 6 and 7.

In the experiments we use the Gaussian kernel defined as $\textbf{K}(\textbf{x}_i, \textbf{x}_j)=\textrm{exp}( -\| \textbf{x}_i - \textbf{x}_j \|^2 / \sigma^2)$, where $\sigma$ is the kernel width. For the graph regularization term we use a fully connected affinity graph with the Gaussian kernel weighting on the edges. To choose the parameter $\sigma$ we perform a simple grid search for the 40 values of $\sigma$ in the range of $[0.1,4]$ with the step size $\Delta \sigma= 0.1$ for data sets Dermatology, Glass, Soybean and Zoo. For the AT\&T face database we perform the grid search in the range  $\sigma = [1000,10000]$ with the step size $\Delta \sigma = 250$. For the Vehicle data set we perform the grid search in the range $\sigma = [10,100]$ with the step size $\Delta \sigma= 10$. At the boundary values of the $\sigma$ intervals the clustering accuracy saturates. For small values of $\sigma$ the similarity of the data points with large distance $\| \textbf{x}_i - \textbf{x}_j \|$ goes to zero as $\textrm{exp}( -\| \textbf{x}_i - \textbf{x}_j \|^2 / \sigma^2) \rightarrow 0$ when $\| \textbf{x}_i - \textbf{x}_j \|^2 / \sigma^2$ is large. Therefore, for small distances, the affinity graph captures the local Euclidean distance and gives a good representation of the manifold structure. For KNSC-Ncut algorithm we used the same grid search to obtain a degree matrix $\textbf{D}^{-1/2}$. 
%
%

For each data set, we measure the average clustering accuracy out of 256 independent runs. We perform a hold-out validation for the parameter $\sigma$, as shown in the Table 4. The hold-out validation consists of randomly splitting each data set into two equally sized parts with the equally distributed cluster membership. The grid search optimization is performed on the first half of the data set, while the second half is used as a hold-out validation where optimized parameters are used. The results of the hold-out validation show robust average clustering accuracy for all three algorithms on all six data sets. 

The sensitivity analysis of the algorithms is performed for the three trade-off parameters $\alpha$, $\mu$ and $\lambda$, plotted in in Fig. 6 and 7. The ratio of the parameters $\mu$ and $\alpha$ is fixed to a constant value in all experiments. The near-orthogonality of the clustering indicator matrix $\textbf{H}$ ($\textbf{Z}$) is preserved during the multiplicative updates, as shown in Fig. 6 and 7. The near-orthogonality of columns is important
for data clustering interpretation. An orthogonal clustering matrix has an interpretation that
each row of $\textbf{H}$  ($\textbf{Z}$) can have only one nonzero element, which
implies that each data object belongs only to 1 cluster. We plot the average orthogonality over 256 runs of the clustering matrix $\textbf{H}$ (KNSC-Rcut) for a wide range of values of the parameter $\mu$ and fixed $\alpha$. The average orthogonality per run is defined as $\sum_{i,i=1}^k (\textbf{HH}^\mathsf{T})_{i,i} / \sum_{i \neq j} (\textbf{HH}^\mathsf{T})_{i,j}$. For a wide range of values of the ratio $\mu / \alpha$ the orthogonality is preserved during the updates. In Fig.  6 we plot the corresponding average clustering accuracy for KNSC-Rcut. When $\mu$ becomes a few order of magnitude larger compared to the reconstruction error term, the objective function effectively becomes the optimization of the orthogonality term. At that point the reconstruction error term loses it significance and the average clustering accuracy starts to drop. In Fig.  6 we plot the clustering accuracy in a wide range of values of the parameter $\mu$. The graph regularization $\lambda$ is fixed to a constant value $\lambda = \alpha$ for simplicity. The average orthogonality is plotted for different values of $\lambda$ and $\mu$ parameters in Fig. 6 and 7. The clustering accuracy is robust for a wide range of $\lambda$, $\lambda = [10^{-4} - 10^2]$, and $\mu$, $\mu = [10^{0} - 10^7]$ throughout the experiments on all six data sets.

%

 
\section{Conclusion}
\label{section4}

In this paper we study subspace clustering from nonlinear orthogonal non-negative matrix factorization perspective. We have constructed a nonlinear orthogonal NMF algorithm and derived three novel clustering algorithms. We have formally shown that the Rcut spectral clustering is equivalent to the nonlinear orthonormal NMF.  The equivalence with the Ncut spectral clustering is obtained by introducing an additional scaling matrix into the nonlinear factorization. Based on this equivalence, we have proposed two kernel-based non-negative spectral clustering methods, KNSC-Ncut and KNSC-Rcut. By incorporating the graph regularization term into the nonlinear NMF framework we have formulated a kernel-based graph-regularized orthogonal non-negative matrix factorization (KOGNMF). To solve the subspace clustering we have derived general kernel-based orthogonal multiplicative updates with complexity $\mathcal{O}(kn^2)$. The monotonic convergence of all three algorithms is proven using an auxiliary function analogous to that used for proving convergence of the Expectation-Maximization algorithm. Experimental results show the effectiveness of our methods compared to state-of-the-art recently proposed NMF-based clustering methods.

  \section*{Acknowledgment}

The authors would like to thank to Mario Lucic and Maria Brbic for proofreading the article. 
The work of DT is funded by the by the 
Croatian Science Foundation IP-2013-11-9623 "Machine learning algorithms for insightful analysis of complex data structures".
The work of IK is funded by Croatian science foundation with the project IP-2016-06-5235 "Structured Decompositions of Empirical Data for Computationally-Assisted Diagnoses of Disease".
The work of NAF s funded by the EU Horizon 2020 SoBigData project under grant agreement No. 654024.


\section*{\textbf{Appendix A}}
{\bf Proof of Theorem 1.}
The factorization $\Phi(\textbf{X})\textbf{D}^{-1/2} = \textbf{WZ}$ can be solved by the following optimization problem
\begin{equation}
\underset{\textbf{Z}, \textbf{W}}{\textrm{min}}\| \Phi(\textbf{X})\textbf{D}^{-1/2} - \textbf{WZ} \|_F^2 \hspace{5pt}s.t. \hspace{5pt} \textbf{Z}\textbf{Z}^\mathsf{T} = \textbf{I},
\label{001}
\end{equation}
where $\textbf{Z}\textbf{Z}^\mathsf{T}  = \textbf{I}$ is the orthogonality constraint which can be included in the optimization implicitly or explicitly via Lagrange multipliers. 
Then objective function can be reformulated as $J(\textbf{Z},\textbf{W})$
\begin{equation}
 \frac{1}{2}\textrm{Tr}\left((\Phi(\textbf{X})\textbf{D}^{-1/2} - \textbf{WZ})^\mathsf{T}(\Phi(\textbf{X})\textbf{D}^{-1/2} - \textbf{WZ})\right)  
 \end{equation}
 \begin{equation}
 = \frac{1}{2}\textrm{Tr}\left((\textbf{D}^{-1/2} \Phi^\mathsf{T}(\textbf{X}) - \textbf{Z}^\mathsf{T}\textbf{W}^\mathsf{T})(\Phi(\textbf{X})\textbf{D}^{-1/2} - \textbf{WZ}) \right)= 
\end{equation}
\begin{equation}
 = \frac{1}{2}\textrm{Tr}\left( \Phi(\textbf{X}) \textbf{D}^{-1}\Phi(\textbf{X})^\mathsf{T} - 2\textbf{W}\textbf{ZD}^{-1/2}\Phi(\textbf{X})^\mathsf{T} + \textbf{WW}^\mathsf{T} \right).
 \label{002}
\end{equation}
The constraint $\textbf{ZZ}^\mathsf{T} = I$ is used in the last equality. Calculating the partial derivative of $J(\textbf{Z},\textbf{W})$ with respect to $\textbf{\textbf{W}}$ and letting it be equal to $0$, it follows

\begin{equation}
 \frac{\partial J(\textbf{Z},\textbf{W})}{\partial \textbf{W}} = - \Phi(\textbf{X}) \textbf{D}^{-1/2} \textbf{Z}^T  + \textbf{W} = 0.
\end{equation}
From here, we have
\begin{equation}
\textbf{W}= \Phi(\textbf{X}) \textbf{D}^{-1/2} \textbf{Z}^\mathsf{T}
\label{004}
\end{equation}
Substituting (\ref{004}) back into (\ref{002}), we  obtain $J(\textbf{Z},\textbf{W}) = $

\begin{equation}
\frac{1}{2}\textrm{Tr}\left( \Phi(\textbf{X}) \textbf{D}^{-1}\Phi(\textbf{X})^\mathsf{T} - 2\Phi(\textbf{X}) \textbf{D}^{-1/2} \textbf{Z}^\mathsf{T}\textbf{Z} \textbf{D}^{-1/2} \Phi(\textbf{X})^\mathsf{T} \right).
\end{equation}
Since  $ \Phi(\textbf{X}) \textbf{D}^{-1}\Phi(\textbf{X})^\mathsf{T} $ is not dependent on $\textbf{Z}$ and $\textbf{W}$, the minimization problem is equivalent to
\begin{equation}
\underset{\textbf{Z}, \textbf{W}}{\textrm{max}}\textrm{Tr}\left(\textbf{Z} \textbf{D}^{-1/2} \Phi(\textbf{X})^\mathsf{T}\Phi(\textbf{X}) \textbf{D}^{-1/2} \textbf{Z}^\mathsf{T} \right) \hspace{5pt}s.t. \hspace{5pt} \textbf{Z}\textbf{Z}^\mathsf{T}  = \textbf{I}.
\label{005}
\end{equation}
For $\textbf{A} = \Phi^\mathsf{T}(\textbf{X})\Phi(\textbf{X})$ the objective function (\ref{005}) is 
\begin{equation}
\underset{\textbf{Z}}{\textrm{max}}\textrm{Tr}\left(\textbf{Z} \textbf{D}^{-1/2} \textbf{A} \textbf{D}^{-1/2} \textbf{Z}^\mathsf{T} \right) \hspace{5pt}s.t. \hspace{5pt} \textbf{Z}\textbf{Z}^\mathsf{T}  = \textbf{I}.
\label{006}
\end{equation}
Note, that the objective function for Ncut spectral clustering 
\begin{equation}
\underset{\textbf{Z}}{\textrm{min}} \textrm{Tr}\left( \textbf{Z} \textbf{L}_{sym} \textbf{Z}^\mathsf{T} \right) \hspace{5pt}s.t. \hspace{5pt} \textbf{Z}\textbf{Z}^\mathsf{T}  = \textbf{I}.
\label{007}
\end{equation}
can easily be transformed to (\ref{005}).
\begin{equation}
\underset{\textbf{Z},\textbf{ZZ}^\mathsf{T} = \textbf{I}}{\textrm{min}} \textrm{Tr}\left( \textbf{Z} \textbf{D}^{-1/2} (\textbf{D}-\textbf{A}) \textbf{D}^{-1/2} \textbf{Z}^\mathsf{T} \right)=
\label{008}
\end{equation}
\begin{equation}
\underset{\textbf{Z}, \textbf{ZZ}^\mathsf{T} = I}{\textrm{min}} \textrm{Tr}\left( \textbf{Z} \textbf{D}^{-1/2}\textbf{D}\textbf{D}^{-1/2} \textbf{Z}^\mathsf{T}  -  \textbf{Z} \textbf{D}^{-1/2}\textbf{A} \textbf{D}^{-1/2} \textbf{Z}^\mathsf{T} \right)=
\label{008}
\end{equation}
and since the term $\textbf{Z} \textbf{D}^{-1/2}\textbf{D}\textbf{D}^{-1/2} \textbf{Z}^\mathsf{T}=\textbf{I}$ due to the orthogonality $\textbf{Z}\textbf{Z}^\mathsf{T}= \textbf{I}$ this is equal to maximization of the second term.
\begin{equation}
\underset{\textbf{Z}, \textbf{Z}\textbf{Z}^\mathsf{T} = \textbf{I}}{\textrm{max}}\textrm{Tr}\left(\textbf{Z} \textbf{D}^{-1/2} \textbf{A} \textbf{D}^{-1/2} \textbf{Z}^\mathsf{T} \right).
\label{006}
\end{equation}
which concludes the Proof. 

\bigskip
{\bf Proof of Theorem 2.} For the Rcut spectral clustering we solve the factorization $\Phi(\textbf{X}) = \textbf{WH}$, with constraint $\textbf{HH}^\mathsf{T} = \textbf{I}$. 
The factorization $\Phi(\textbf{X}) = \textbf{WH}$ can be solved by the optimization problem
\begin{equation}
\underset{\textbf{H}, \textbf{W}, \textbf{HH}^\mathsf{T} = \textbf{I}}{\textrm{min}}\| \Phi(\textbf{X}) - \textbf{WH}\|_F^2, 
\label{007}
\end{equation}
where $\textbf{HH}^\mathsf{T} = \textbf{I}$ is the orthogonality constraint which can be included in the optimization implicitly or explicitly. The objective function (\ref{007}) can be reformulated as
\begin{equation}
J(\textbf{H},\textbf{W}) = \frac{1}{2}\textrm{Tr}\left((\Phi(\textbf{X}) - \textbf{WH})^\mathsf{T}(\Phi(\textbf{X}) - \textbf{WH})\right) = 
\end{equation}
\begin{equation}
 = \frac{1}{2}\textrm{Tr}\left((\Phi^\mathsf{T}(\textbf{X}) - \textbf{H}^\mathsf{T}\textbf{W}^\mathsf{T})(\Phi(\textbf{X}) - \textbf{WH}) \right)= 
\end{equation}
\begin{equation}
 = \frac{1}{2}\textrm{Tr}\left( \Phi(\textbf{X})^\mathsf{T}\Phi(\textbf{X}) - 2\Phi(\textbf{X})^\mathsf{T}\textbf{WH} + \textbf{W}^\mathsf{T}\textbf{W} \right).
 \label{008}
\end{equation}
The constraint $\textbf{HH}^\mathsf{T} =\textbf{I}$ is used in the last equality. Calculating the partial derivative of $J(\textbf{H},\textbf{W})$ with respect to $\textbf{W}$ and letting it be equal to $0$, it follows

\begin{equation}
 \frac{\partial J(\textbf{H},\textbf{W})}{\partial \textbf{W}} = - \Phi(\textbf{X}) \textbf{H}^\mathsf{T} + \textbf{W}= 0.
\end{equation}
From here, we have
\begin{equation}
\textbf{W} = \Phi(\textbf{X}) \textbf{H}^\mathsf{T}.
\label{009}
\end{equation}
Substituting (\ref{009}) back into (\ref{008}), we obtain 
\begin{equation}
J(\textbf{H}) = \frac{1}{2}\textrm{Tr}\left( \Phi(\textbf{X})^\mathsf{T} \Phi(\textbf{X}) - \Phi(\textbf{X})^\mathsf{T} \Phi(\textbf{X}) \textbf{H}^\mathsf{T} \textbf{H}  \right).
\end{equation}
Since the first term is constant, not dependent on $\textbf{H}$ and $\textbf{W}$, the minimization problem is equivalent to 
\begin{equation}
\underset{\textbf{H}, \textbf{W}, \textbf{HH}^\mathsf{T} = I}{\textrm{max}}\textrm{Tr}\left(\textbf{H} \Phi(\textbf{X})^\mathsf{T}\Phi(\textbf{X}) \textbf{H}^\mathsf{T} \right).
\label{010}
\end{equation}
For $\textbf{A} = \Phi^\mathsf{T}(\textbf{X})\Phi(\textbf{X})$ the objective function (\ref{010}) is the same as objective function (\ref{006}) for the relaxed Rcut spectral clustering. To see why, we start from the objective function of Rcut and come to the relaxed Rcut optimization function \cite{16}:
\begin{equation}
\underset{\textbf{H}, \textbf{HH}^\mathsf{T} = \textbf{I}}{\textrm{min}}\textrm{Tr}\left(\textbf{H}\textbf{L}\textbf{H}^\mathsf{T} \right) = \underset{\textbf{H}, \textbf{HH}^\mathsf{T} = \textbf{I}}{\textrm{min}}\textrm{Tr}\left(\textbf{H} \textbf{D} \textbf{H}^\mathsf{T} - \textbf{H} \textbf{A} \textbf{H}^\mathsf{T} \right).
\label{011}
\end{equation}
Now, the substitution is made $\textbf{Q}= \textbf{H} \textbf{D}^{1/2}$ which implies $\textbf{H}=\textbf{Q}\textbf{D}^{-1/2}$, $\textbf{HH}^\mathsf{T} = \textbf{Q} \textbf{D}^{-1} \textbf{Q}^\mathsf{T}$and the objective function can we written as:
\begin{equation*}
\underset{\textbf{Q}, \textbf{Q} \textbf{D}^{-1} \textbf{Q}^\mathsf{T} = \textbf{I}}{\textrm{min}}\textrm{Tr}\left(\textbf{QD}^{-1/2} \textbf{DD}^{-1/2}\textbf{Q}^\mathsf{T} - \textbf{QD}^{-1/2} \textbf{A}\textbf{D}^{-1/2}\textbf{Q}^\mathsf{T} \right) 
\end{equation*}
\begin{equation}
= \underset{\textbf{Q}, \textbf{QD}^{-1} \textbf{Q}^\mathsf{T} = \textbf{I}}{\textrm{min}}\textrm{Tr}\left(\textbf{QQ}^\mathsf{T} - \textbf{QD}^{-1/2}\textbf{AD}^{-1/2}\textbf{Q}^\mathsf{T} \right).
\label{012}
\end{equation}
The expression (\ref{012}) is equivalent to
\begin{equation}
 \underset{\textbf{Q}, \textbf{Q}\textbf{D}^{-1} \textbf{Q}^\mathsf{T} = \textbf{I}}{\textrm{max}} \textrm{Tr}\left(\textbf{QD}^{-1/2} \textbf{AD}^{-1/2}\textbf{Q}^\mathsf{T} \right) \hspace{5pt}s.t. \hspace{5pt} \textbf{QQ}^\mathsf{T} = \textbf{I}
\label{013}
\end{equation}
Next, we release the orthonormality constraint $\textbf{QQ}^\mathsf{T} = \textbf{I}$. The relaxation is justified by the fact that the rows of $\textbf{Q}$ are orthogonal to each other since $\textbf{QD}^{-1}\textbf{Q}^\mathsf{T} = \textbf{I}$. 
\begin{equation}
 \underset{\textbf{Q}, \textbf{QD}^{-1} \textbf{Q}^\mathsf{T} = I}{\textrm{max}}\textrm{Tr}\left(\textbf{QD}^{-1/2}\textbf{AD}^{-1/2}\textbf{Q}^\mathsf{T} \right) 
\label{014}
\end{equation}
and by substitution $\textbf{Q}= \textbf{HD}^{1/2}$ this becomes:
\begin{equation}
 \underset{\textbf{H}, \textbf{HH}^\mathsf{T} = \textbf{I}}{\textrm{max}}\textrm{Tr}\left(\textbf{HAH} \right) 
\label{014}
\end{equation}
which is equal to objective function of (\ref{010}), which concludes the Proof. 

\section*{Appendix B}
\textbf{Proof 3. The convergence analysis of the proposed algorithms.}

We now show the algorithm KOGNMF converges to a feasible solution. We use the auxiliary function approach, following \cite{19,50}. The convergence of KNSC-Ncut and KNSC-Rcut can be proven in a similar way. 

The objective function of KOGNMF (36) is non-increasing under the alternative iterative updating rules in (37) and (38).

\textbf{Definition.} $A(h,h^{\prime})$ is an auxiliary function for $B(h)$ when the following conditions are satisfied:

\begin{equation}
A(h,h^{\prime}) \geq B(h), \hspace{1cm} A(h,h) = B(h).
\end{equation}
The auxiliary function is useful because of the following lemma: \\
\textbf{Lemma 1.} If $A$ is an auxiliary function of  $B$, then $B$ is non-increasing under the updating formula

\begin{equation}
h^{(t+1)} = \underset{h}{\textrm{arg min }} A(h,h^{(t)})
\end{equation}
the function $B$ is non-increasing. 

\textbf{Proof.} $B(h^{(t+1)}) \leq A(h^{(t+1)}, h^{(t)}) \leq A(h^{(t)}, h^{(t)}) = B(h^{(t)}).$

We now rewrite the objective function $\mathcal{L}$ of KOGNMF in Eq. (36) as follows

\begin{equation*}
\mathcal{L} = \alpha\| \Phi(\textbf{X}) - \Phi(\textbf{X})\textbf{FH} \|^2_F + \lambda \textrm{Tr}(\textbf{HLH}^\mathsf{T}) + \mu \|\textbf{HH}^\mathsf{T}-\textbf{I}_k\|^2_F
\end{equation*}
\begin{equation}
 = \alpha \sum_{i=1}^D \sum_{j=1}^n\left(\Phi(x)_{ij} - \sum_{l = 1}^{k}w_{il}h_{lj}\right)^2 + \lambda \sum_{m=1}^k \sum_{j=1}^n \sum_{l=1}^n h_{mj} L_{jl} h_{lm} + 
 \mu \sum_{i=1}^D \sum_{j=1}^n \left(\sum_{l = 1}^{n} h_{ik}h_{kj} -  \delta_{ij} \right)^2 
\end{equation}

Considering any element $h_{ab}$ in $\textbf{H}$, we use $B_{ab}$ to denote the part of $\mathcal{L}$ relevant to $h_{ab}$. Then it follows

\begin{equation}
\dot{B}_{ab} \equiv \left(\frac{\partial \mathcal{L}}{\partial \textbf{H}} \right)_{ab} =  \left( 2\alpha\textbf{F}^\mathsf{T}\textbf{K}\textbf{F}\textbf{H} - 2\alpha\textbf{F}^\mathsf{T} \textbf{K} + 2\lambda\textbf{HL} + 4\mu \textbf{H}(\textbf{H}^\mathsf{T}\textbf{H}-\textbf{I}) \right)_{ab}
\end{equation}

Since multiplicative update rules are element-wise, we have to show that each $B_{ab}$ is non-increasing under the update step given in Eq. (37).

\textbf{Lemma 2.} Function 
\begin{equation}
A(h,h_{ab}^{(t)}) = B(h_{ab}^{(t)}) + \dot{B}_{ab} (h_{ab}^{(t)})(h - h_{ab}^{(t)}) + \frac{(2\alpha\textbf{F}^\mathsf{T} \textbf{K} \textbf{FH} + 2\lambda \textbf{HD} )_{ab}}{h^t_{ab}}(h - h^t_{ab})^2. 
\label{aux}
\end{equation}
is an auxiliary function for $B_{ab}$, when $\mu=0$.

\textbf{Proof.} By the above equation, we have $A(h,h) = B_{ab}(h)$, so we only need to show that $A(h,h^t_{ab}) \geq B_{ab}(h).$ To this end, we compare the auxiliary function given in Eq. (\ref{aux}) with the Taylor expansion of $B_{ab}(h)$.
\begin{equation}
\ddot{B}_{ab} \equiv \left(\frac{\partial^2 \mathcal{L}}{\partial \textbf{H}^2}\right)_{ab} = \left( 2\alpha \textbf{F}^\mathsf{T} \textbf{K} \textbf{F} + 2\lambda \textbf{L} \right)_{ab}
\end{equation}
\begin{equation}
B_{ab}(h) = B_{ab}(h^{(t)}_{ab}) + \dot{B}_{ab}(h-h^{(t)}_{ab})+[\alpha\textbf{F}^\mathsf{T} \textbf{K} \textbf{F} + \lambda \textbf{L}]_{ab}(h - h^{(t)}_{ab})^2 
\end{equation}
to find that $A(h,h^t_{ab}) \geq B_{ab}(h)$ is equivalent to

\begin{equation}
\frac{\alpha(\textbf{F}^\mathsf{T} \textbf{K} \textbf{FH})_{ab} + \lambda (\textbf{HD})_{ab} }{h^t_{ab}} \geq (\alpha\textbf{F}^\mathsf{T} \textbf{K} \textbf{F} + \lambda \textbf{L})_{ab}
\end{equation}


%
%

\begin{equation}
(\textbf{F}^\mathsf{T} \textbf{K} \textbf{FH})_{ab} = \sum_{l=1}^{k} (\textbf{F}^\mathsf{T} \textbf{K} \textbf{F})_{al} h^t_{lb} \geq (\textbf{F}^\mathsf{T} \textbf{K} \textbf{F})_{aa} h^t_{ab}
\end{equation}

\begin{equation}
(\textbf{HD})_{ab} = \sum_{l=1}^{n} h^t_{al} \textbf{D}_{lb} \geq h^t_{ab} \textbf{D}_{bb} \geq h^t_{ab} (\textbf{D} - \textbf{A})_{bb}
\end{equation}


In summary, we have the following inequality

\begin{equation}
\frac{(\alpha \textbf{F}^\mathsf{T} \mathbf{K} \textbf{FH} +  \lambda \textbf{HD} )_{ab}}{h^t_{ab}} \geq \frac{1}{2} \ddot{B}_{ab}
\end{equation}

Then the inequality $A(h,h^t_{ab}) \geq B_{ab}(h)$ is satisfied, and the Lemma is proven.

From Lemma 2, we know that $A(h,h^t_{ab})$ is an auxiliary function of $B_{ab}(h_{ab})$.  We can now demonstrate the convergence of the update rules given in Eqs. (37).

\begin{equation}
h^{t+1} = \underset{h}{\textrm{arg min}} A(h,h^{(t)})
\end{equation}


\begin{equation}
h^{t+1}_{ab} = h^{t}_{ab}\frac{   (\alpha \textbf{F}^\mathsf{T}\mathbf{K} +  \lambda \textbf{HA})_{ab}     }
{ ( \alpha \textbf{F}^\mathsf{T} \mathbf{K} \textbf{FH} +  \lambda \textbf{HD} )_{ab} }
\end{equation}

So the updating rule for $H$ is as follows:

\begin{equation}
H_{ab} \leftarrow H_{ab} \frac{(\alpha \textbf{F}^\mathsf{T}\mathbf{K} +  \lambda \textbf{HA})_{ab}}{(\alpha \textbf{F}^\mathsf{T} \mathbf{K} \textbf{FH } + \lambda \textbf{HD})_{ab}}
\end{equation}


Similarly, for $\mu>0$, we use the following auxiliary function $A(h,h^t_{ab}) =$

\begin{equation}
A(h,h_{ab}^{(t)}) = B(h_{ab}^{(t)}) + \dot{B}_{ab} (h_{ab}^{(t)})(h - h_{ab}^{(t)}) + \frac{\alpha(\textbf{F}^\mathsf{T} \textbf{K} \textbf{FH})_{ab} + \lambda (\textbf{HD})_{ab} + \mu (\textbf{HH}^\mathsf{T}\textbf{H} )_{ab}}{h^t_{ab}}(h - h^t_{ab})^2. 
\end{equation}
and by using this:
\begin{equation}
 (\textbf{HH}^\mathsf{T} \textbf{H})_{ab} =  \sum_{l=1}^{n} h^t_{al}  (\textbf{H}^\mathsf{T} \textbf{H} )_{lb} 
 \geq
  h^t_{ab} (\textbf{H}^\mathsf{T}\textbf{H})_{bb}
\end{equation}
we obtain the following inequality
\begin{equation}
\frac{\alpha(\textbf{F}^\mathsf{T} \textbf{K} \textbf{FH})_{ab} + \lambda (\textbf{HD})_{ab} + \mu (\textbf{HH}^\mathsf{T}\textbf{H})_{ab}}{h^t_{ab}} \geq (\alpha\textbf{F}^\mathsf{T} \textbf{K} \textbf{F} + \mu \textbf{H}^\mathsf{T} \textbf{H} + \lambda \textbf{L})_{ab}
\end{equation}
which is used to prove that (86) is an auxiliary function of (74).
Finally, we get the update rule
\begin{equation}
H_{ab} \leftarrow H_{ab} \frac{(\alpha \textbf{F}^\mathsf{T}\mathbf{K} + 2\mu \textbf{H} + \lambda \textbf{HA})_{ab}}{(\alpha \textbf{F}^\mathsf{T} \mathbf{K} \textbf{FH }+ 2\mu  \textbf{HH}^\mathsf{T} \textbf{H} + \lambda \textbf{HD})_{ab}}.
\end{equation}

 \vspace{0.5cm}

The proof of the convergence for the $\textbf{F}$ update rule (38) can be derived by following proposition 8 from \cite{50}. The auxiliary function for our objective function $\mathcal{L}(\textbf{F})$ (39) as a function of $\textbf{F}$ is: 
\begin{equation}
A(\textbf{F},\textbf{F}^{'}) = -\sum_{i,k} 2(\textbf{K}\textbf{H}^\mathsf{T})_{i,k} \textbf{F}_{i,k}^{'}(1+log\frac{\textbf{F}_{ik}}{\textbf{F}^{'}_{ik}}) + \sum_{i,k} \frac{(\mathbf{K}  \textbf{F}^{'} \textbf{HH}^\mathsf{T})_{i,k} (\textbf{F}_{i,k})^2}{\textbf{F}^{'}_{i,k}},
\end{equation}
The proof that this is an auxiliary function of $\mathcal{L}(\textbf{F})$ (39) is given in \cite{50}, with the change in notation $\textbf{F}=\textbf{W}$, $\textbf{H}=\textbf{G}^\mathsf{T}$ and $\Phi(\textbf{X}) = \textbf{X}$. \\
This auxiliary function is a convex function of $F$ and it's global minimum can be derived with the following update rule:
\begin{equation}
\label{eq:FupdateGr}
F_{ab} \leftarrow F_{ab} \frac{(\mathbf{K} \textbf{H}^\mathsf{T})_{ab}}{(\textbf{K} \textbf{FHH}^\mathsf{T})_{ab}}.
\end{equation}

\bibliographystyle{IEEEtran}
\bibliography{NMF_SC_references}{}

\end{document}